\PassOptionsToPackage{table}{xcolor}
\documentclass[sn-mathphys]{sn-jnl}

\usepackage[utf8]{inputenc} 
\usepackage[T1]{fontenc}    
\usepackage{hyperref}       
\usepackage{url}            
\usepackage{booktabs}       
\usepackage{amsfonts}       
\usepackage{nicefrac}       
\usepackage{lipsum}
\usepackage{graphicx}
\usepackage[english]{babel}
\usepackage[table]{xcolor}
\usepackage{float}
\usepackage{amsthm}
\usepackage{amsmath}
\usepackage{multirow}
\usepackage[none]{hyphenat}
\usepackage{rotating}
\usepackage{placeins}
\usepackage{ulem}

\jyear{2021}%

\theoremstyle{thmstyleone}%
\theoremstyle{thmstyletwo}%

\theoremstyle{thmstylethree}%
\newtheorem{definition}{Definition}%
\newtheorem{notation}{Notation}%
\begin{document}

\title[\footnotesize A feature selection method based on Shapley values robust for concept shift in regression]{A feature selection method based on Shapley values robust for concept shift in regression}

\author*[1,2]{\fnm{Carlos} \sur{Sebastián}}\email{carlos.sebastian@fortiaenergia.es}

\author[3,4]{\fnm{Carlos E.} \sur{González-Guillén}}\email{carlos.gguillen@upm.es}

\affil*[1]{\orgname{Fortia Energía}, \orgaddress{\street{Calle de Gregorio Benítez}, \city{Madrid}, \postcode{28043}, \country{Spain}}}

\affil[2]{\orgname{Universidad Politécnica de Madrid}, \orgaddress{\city{Madrid}, \country{Spain}}}

\affil[3]{\orgdiv{Departamento de Matemática Aplicada a la Ingeniería Industrial, Escuela Técnica Superior de Ingenieros Industriales}, \orgname{Universidad Politécnica de Madrid}, \orgaddress{\street{Calle de José Gutiérrez Abascal}, \city{Madrid}, \postcode{28006}, \country{Spain}}}

\affil[4]{\orgname{Instituto de Ciencias Matemáticas (CSIC-UAM-UC3M-UCM)}, \orgaddress{\street{Calle Nicolás Cabrera}, \city{Madrid}, \postcode{28049}, \country{Spain}}}

\abstract{Feature selection is one of the most relevant processes in any methodology for creating a statistical learning model. Usually, existing algorithms establish some criterion to select the most influential variables, discarding those that do not contribute to the model with any relevant information. This methodology makes sense in a static situation where the joint distribution of the data does not vary over time. However, when dealing with real data, it is common to encounter the problem of the dataset shift and, specifically, changes in the relationships between variables (concept shift). In this case, the influence of a variable cannot be the only indicator of its quality as a regressor of the model, since the relationship learned in the training phase may not correspond to the current situation. In tackling this problem, our approach establishes a direct relationship between the Shapley values and prediction errors, operating at a more local level to effectively detect the individual biases introduced by each variable. The proposed methodology is evaluated through various examples, including synthetic scenarios mimicking sudden and incremental shift situations, as well as two real-world cases characterized by concept shifts. Additionally, we perform three analyses of standard situations to assess the algorithm's robustness in the absence of shifts. The results demonstrate that our proposed algorithm significantly outperforms state-of-the-art feature selection methods in concept shift scenarios, while matching the performance of existing methodologies in static situations.}


\maketitle

\section{Introduction}

Artificial Intelligence and Machine Learning in particular are becoming increasingly valuable nowadays. The understanding of the vast amount of data available being generated every day in a connected world, offers many possibilities like Natural Language Processing systems, recommendation engines, medical diagnosis or forecasting models. Thus, the use of historical data to build models capable of capturing relationships between different variables is becoming more and more frequent, either for explanatory or predictive purposes. However, different issues may arise when modelling a problem: the tradeoff between the explainability and accuracy of different algorithms, computational complexity... A common problem to all fields is the selection of the correct information in high-dimensional contexts. Failure to perform this task properly can lead to problems of overfitting, that is, of poor generalisation.\\

Dimension reduction is generally performed in two ways: feature extraction (FE) or feature selection (FS). FE consists of projecting the feature space in a high dimensional space to a smaller one, while FS chooses a subset of the original features \citep{venkatesh2019review}. In a large number of applications, transforming the original variables can make it challenging to analyze the results. This modification can be problematic when we need the outcomes to be interpretable, as the intuition behind the original features is lost. That is why, in such cases, feature selection is usually preferred \citep{li2017feature}. Moreover, even when the dimension of the data is not so high, it is important to follow the principle of parsimony or Ockham's razor, as the models built will be more explainable and, in case they are used as predictive models, more general and robust.\\

When fitting a model with real data, a very common problem is the so-called dataset shift. This occurs when the statistical properties of the target variable change over time \citep{widmer1996learning}. This shift may occur for seemingly arbitrary reasons, but it may also be due to changes in the explanatory variables. If there is not sufficiently long history of data available since the change occurred and the affected variables are in the explanatory variables of the model, the relationship learned by the model through these features may be erroneous.\\

Feature selection algorithms do not detect this kind of problem, as they usually take as reference a measure of the degree of importance of the variable to determine whether it is relevant or not. However, they do not observe whether the effect that each variable has on the predictions is the desired one, regardless of the magnitude of the effect.\\

We introduce a new method for variable selection in regression problems that is robust to the presence of characteristics whose behaviour has changed over time and have an undesirable effect on predictions in the current context.  This selection is done using Shapley values, as previously considered in \cite{marcilio2020explanations}, \cite{eoghan_keany_2020_4247618}, \cite{Shapicant} or \cite{verhaeghe2022powershap}. While all these works consider a global treatment of each feature to asses their importance, our method relates Shapley values to the errors of the predictions in order to pay special attention to the local effects of each variable for each prediction.\\

It is important to note that we do not propose a method to detect or characterise a dataset shift, but rather that in the event of any variable undergoing some sort of shift that negatively influences the performance of the model, this variable is a candidate for elimination, regardless of the overall level of influence it has on the behaviour of the model.\\

The rest of the paper is organised as follows: Section \ref{sec:feature_selection} describes the types of feature selection algorithms. Section \ref{sec:Shapley} introduces Shapley values in the context of Machine Learning, discussing their inclusion in different variable selection algorithms. Section \ref{sec:shift} introduces the dataset shift problem and its importance in the production of predictive models in real applications. In Section \ref{sec:algoritmo} the new proposed algorithm is explained in detail, distinguishing the results obtained in Section \ref{sec:experimentos}. Finally, the work is concluded in Section \ref{sec:conclusiones}, opening also new possible lines of research.\\

\section{Previous work}\label{sec:feature_selection}

Feature selection is a complicated task when there is no help from experts in the field. The selection in such cases has to be made directly from the data and optionally by using statistical models. Establishing the state of the art in variable selection methods is a difficult task, as it depends considerably on the task to be performed (classification or regression) and the dataset employed. However, there is a clear classification of the different types of existing methods: filters, wrappers and embedded methods.
\paragraph*{Filters}Filters make a selection of features using only the relationships that exist between the data, they do not rely on the use of any model. Thus, some criterion is used to establish a ranking among the variables and those that exceed a certain threshold in this ranking are chosen as relevant. Examples of different criteria are: correlation with the target variable, information gain, Fisher score, variance threshold or the chi-square test \citep{venkatesh2019review}. Since they do not rely on a model, they tend to be computationally very efficient methods. However, they require making assumptions about the data that are not always met, leading to a selection that is not necessarily optimal under the criteria used.

\paragraph*{Wrappers}Wrapper methods, popularised by \cite{kohavi1997wrappers}, are those that interact with a predefined model to assess the quality of the selected feature set. The methodology is divided into two steps: finding a feature set and assessing the quality of the feature set \citep{li2017feature}. Three common strategies when proposing a dataset are forward selection, backward selection and stepwise selection \citep{colaco2019review}. The use of metaheuristics, typically bio-inspired \citep{diao2015nature}, is also common. There are methods that were specifically designed for certain models, such as the Boruta method \citep{kursa2010feature} for Random Forest models. The constant interaction with a model makes them computationally more expensive than filter methods, although the results are usually better \citep{venkatesh2019review}.

\paragraph*{Embedded methods}Embedded methods are those that incorporate variable selection into the learning phase of the model. In this way they are computationally efficient and take advantage of the benefits of interacting with a model. The most common are those that add some kind of regularisation to the learning process, so that the coefficients of certain features are forced to be zero or close to zero \citep{li2017feature}. Although there are complex methods, it is very common to use Lasso regression to select variables, obtaining more than acceptable results on many occasions, such as in time series forecasting \citep{petropoulos2022forecasting}.

\subsection{Shapley values in the context of feature selection}\label{sec:Shapley}

Shapley values \citep{Shapley} are a game theory concept to explain the importance of an individual player in a collaborative team. The idea behind the notion is based on distributing the total gain among the players according to the relative importance of their contributions. Specifically, the Shapley value for player $i$ is defined as the average marginal contribution over all possible coalitions. This is $$ \phi_i = \sum_{ \mathcal{S} \subseteq \mathcal{N}\setminus\{i\}} \frac{\vert \mathcal{S} \vert ! \ (\vert \mathcal{N} \vert - \vert \mathcal{S} \vert - 1) !}{\vert \mathcal{N} \vert !} \left( v(\mathcal{S} \cup \{i\}) - v(\mathcal{S}) \right)$$ where $\mathcal{N}$ is the set of players and $v: \mathcal{P}(\mathcal{N}) \longrightarrow \mathbb{R}$ a function that returns the gain of a coalition.\\

This idea can be transferred to the explainability of Machine Learning models \citep{vstrumbelj2014explaining}. Specifically, the game would be the task of predicting an instance of the dataset, the gain would be the difference of the prediction with the average prediction of all instances, and the players would be the values of the different variables that collaborate to obtain the prediction. In this way, Shapley values allow us to measure the influence that each variable has on a prediction, distinguishing the characteristics that have a higher or lower impact on the predictions made by a model. However, for high dimensional data the calculation of the Shapley values in an exact way is not feasible, as the computational complexity is $\mathcal{O}(2^{\vert \mathcal{N} \vert })$.\\

The use of Shapley values as a local model-agnostic technique for the explainability of Machine Learning models became popular with the introduction of SHAP (SHapley Additive exPlanations) \citep{lundberg2017unified} and in particular with the appearance of TreeSHAP \citep{lundberg2018consistent}, which allows the efficient calculation of an approximation of Shapley values for models based on decision trees. Although this approximation presents problems estimating non-zero Shapley values in features that are not relevant but have a high correlation with variables that are influential \citep{sundararajan2020many, janzing2020feature}, it works well in practice.\\

For other types of models, such as neural networks, there are techniques such as the one presented in \cite{castro2009polynomial} or \cite{ancona2019explaining} that allow the calculation of an approximation of the Shapley value in polynomial time. For instance, the first approximation is based on random sampling from the equivalent definition of Shapley value $$ \phi_i = \frac{1}{\vert \Pi(\mathcal{N}) \vert} \sum_{\pi \in \Pi(\mathcal{N})}\left( v(\mathcal{P}_i^\pi \cup \{i\}) - v(\mathcal{P}_i^\pi) \right)$$ where $\Pi(\mathcal{N})$ denotes the set of permutations of $\mathcal{N}$ and $\mathcal{P}_i^\pi = \{j \in \mathcal{N}\, : \, \pi(j) < \pi(i)\}$ is the so-called predecessor set of the player $i$, that takes the position $\pi(i)$ in the permutation $\pi \in \Pi(\mathcal{N})$.\\

Taking into account that the Shapley value can be used to measure the influence of a characteristic on a prediction, an overall measure of the influence of a variable can be defined by taking the mean of the absolute values of the Shapley values of every observation for each feature. This measure can be used to filter by a minimum overall influence or to establish a ranking and select the $k$ most important variables \citep{marcilio2020explanations}. In practice this method works well, although the idea of using Shapley values within the variable selection field can be refined considerably.\\

A first idea is the variation of the Boruta \citep{kursa2010feature} method, Boruta-Shap \citep{eoghan_keany_2020_4247618}. Boruta-Shap, like Boruta, is based on the use of shadow variables, which are copies of the original variables with the values randomly permuted. The general idea is that a variable is relevant if its importance measure is greater than that of the best shadow variable. Boruta-Shap modifies the algorithm by using the previously described global influence as a measure of importance and making technical modifications to speed up the process, establishing that not only better feature sets are obtained than through the original algorithm, but also in less time.\\

Another modification is Shapicant \citep{Shapicant}, which is inspired by the Permutation Importance (PIMP) method, \citep{altmann2010permutation}. In PIMP, a model is trained on the original data and an importance measure is obtained for each variable. Then, a permutation of the target variable is made and the model is re-trained to obtain another importance score for each of the variables. The process is repeated several times and if a variable is significantly more relevant on average over the original dataset than over the randomised set, then it is considered to be relevant. Shapicant uses the Shapley values as a measure of importance, but separating positive and negative values.\\

A new original method was proposed by \cite{verhaeghe2022powershap}, Powershap. This algorithm is divided into two phases: in the \textit{Explain} component, a uniform random variable is added to the dataset, a given model is trained and the overall influence measure is calculated based on Shapley values for each of the variables, including the random variable. This procedure is repeated for a fixed number of iterations but varying the random variable associated with the model to obtain different results. In the \textit{Core} part, the performance of all the features is compared with that of the random variable and the most important variables are determined through a hypothesis test. Furthermore, it proposes an automatic method to optimise the hyperparameter associated with the number of iterations of the first component while keeping fixed the threshold for the p-value.\\

It is noteworthy that all these algorithms presented include the Shapley values through the mean of the absolute values for each variable as a measure of global influence.\\

\subsection{Dataset shift}\label{sec:shift}

The concept of dataset shift was introduced in \cite{quinonero2008dataset}, where it is described as a phenomenon in which the joint distribution of the explanatory variables and the target variable is different between the training set and the test set used in the creation of a statistical learning model. More formally, given a time period $\left[0, \, t\right]$, a set of samples denoted $S_{0, t} = \left\lbrace d_0, \dots, d_t \right\rbrace$, where $d_i = (X_i, y_i)$ with $X_i$ the vector of explanatory variables and $y_i$ the target variable. Let $F_{0, t}(X, y)$ be the distribution following $S_{0, t}$, analogously denote  $S_{t+1, \infty}$ and $F_{t+1, \infty}(X, y)$ for a time period $\left[t+1, \, \infty\right)$. A dataset shift is said to occur at time $t+1$ if $F_{0, t}(X, y) \neq F_{t+1, \infty}(X, y)$, i.e.  $\exists \; t \, : \, P_t(X, y)  \neq P_{t+1}(X, y)$ \citep{lu2018learning}.\\

Independently of the model used, a typical data analysis scheme assumes that the distribution of the data is static for the model to be valid. If there is a variation in the distribution, that change must be modelled. Such changes are very common in real problems, such as economic, political, social, regulatory, etc., reasons that affect the behaviour of many phenomena. This is the reason why the dataset shift problem must be addressed.\\

\begin{figure}[h]
\centering
\includegraphics[scale=0.5]{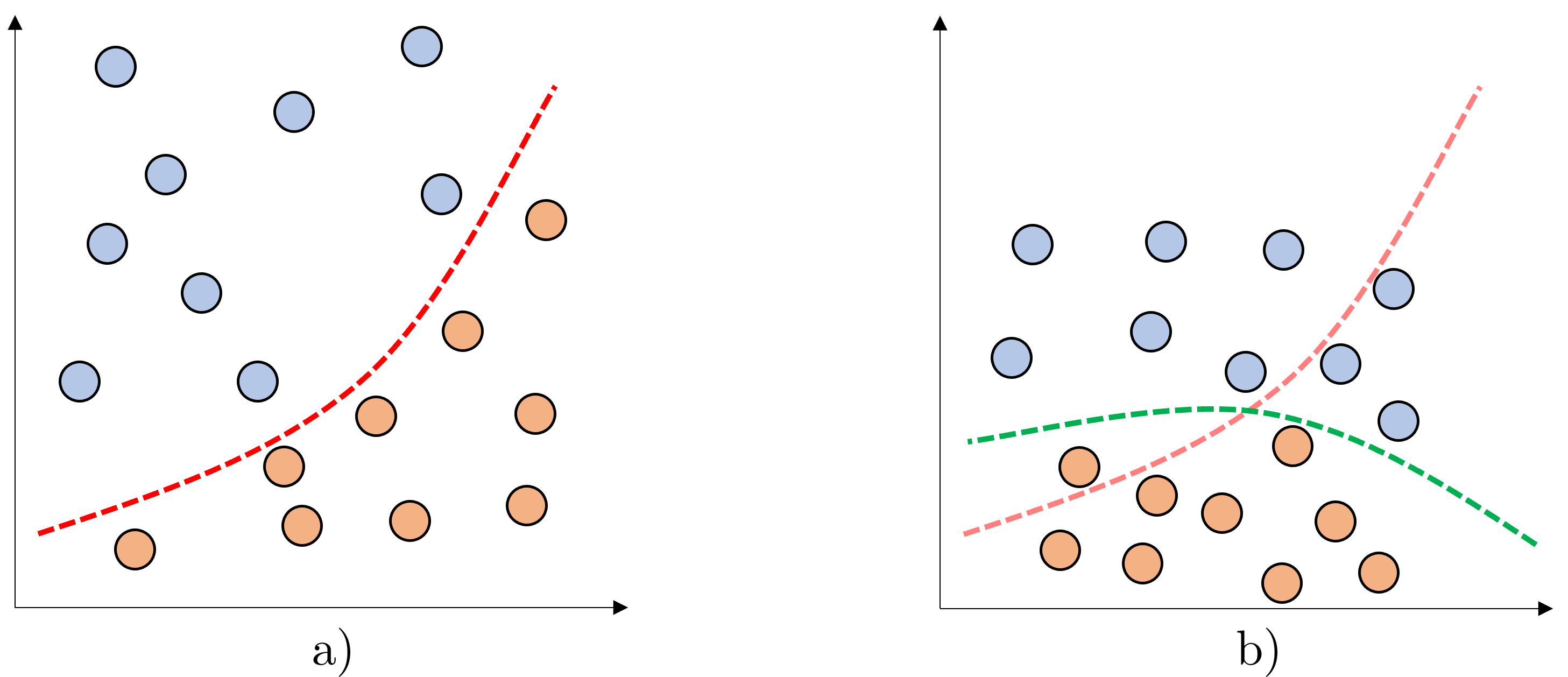}
\caption{In \textbf{a)} the learned decision frontier is observed with respect to the training data. In \textbf{b)} it can be seen that the relationship between the variables has changed and that the decision frontier learned in training is not valid for the test data.}
\label{fig:concept_shift}
\end{figure}

Let $t+1$ be the instant at which the dataset shift occurs, there are three possible reasons why the joint probability distribution is different \citep{moreno2012unifying}:
\begin{enumerate}
\item \textit{Covariate shift} which is probably the most studied type of shift and happens when $P_t(y \vert X) = P_{t+1}(y \vert X)$ but $P_t(X) \neq P_{t+1}(X)$.
\item \textit{Prior probability shift}, when $P_t(X \vert y) = P_{t+1}(X \vert y)$ but $P_t(y) \neq P_{t+1}(y)$
\item \textit{Concept shift}, which occurs when the relationship between the explanatory and target variables change, that is, when $P_t(y \vert X) \neq P_{t+1}(y \vert X)$ but $P_t(X) = P_{t+1}(X)$ or when $P_t(X \vert y) \neq P_{t+1}(X \vert y)$ but $P_t(y) = P_{t+1}(y)$. It is the most complex type of shift. An example in a classification problem is shown in Figure \ref{fig:concept_shift}.
\end{enumerate}

Much of the related research in this field focuses on shift detection (whether shift occurs or not), on understanding why it occurs (when, how and where) and on shift adaptation (reacting to change). It is also often treated from the perspective of classification problems, while the field of regression has not been explored in great depth \citep{lima2022learning}. Most strategies designed to react to the presence of the shift are based on retraining the models with more current data or with data similar to those occurring in the current context, although other strategies are possible \citep{lu2018learning}. The implementation of such strategies can be complicated and requires constant maintenance. In addition, many algorithms require a large amount of data to perform satisfactorily, which severely limits the use of data related to the current paradigm.\\

In the particular case of a concept shift, where the relationship between the explanatory variables and the target variable changes, the situation where only one group of variables is responsible for the shift may arise. If the elimination of this group of variables still allows the construction of a good statistical model, the full use of the dataset and a much simpler model maintenance can be achieved, as the shift will no longer be present.\\

\section{A new feature selection algorithm for regression problems}\label{sec:algoritmo}

Generally, feature selection algorithms quantify the importance of a variable and, based on some criterion, determine a minimum threshold of importance to be considered. However, they do not take into account the effect that a variable has, i.e. a feature may have a large impact on the decisions a model makes, but it does not necessarily need to have the right influence. This should not be the case in a standard static situation where the learned behaviour directly corresponds to the actual behaviour, but under a concept shift setting this situation is quite common.\\

Here, we propose a new variable selection algorithm that is robust to these situations. It is able to detect the features that cause the shift in case they are among the possible explanatory variables and, in situations where there is no shift, performs similar or better than the state of the art.\\ 

\subsection{The intuitive idea}

To achieve the goal of detecting the features that do not have the right influence on the predictions, we will analyze at a more local level the effect that each variable has on each prediction. We are going to estimate this effect using the Shapley value of the feature over the prediction. Then, depending on the error we are making in each one of the observations (over or under predicted), we are going to consider it a positive or negative effect. For instance, if an observation is under predicted, then a feature that decreases the prediction value is not ideal, because it is contributing to making that error higher. After that, we are going to compute if a feature has more positive effects or negative effects and in case the second is higher, we are going to assign this feature a ``negative influence'' and it will be considered for removal in an iterative process.\\

At a high level, the algorithm classifies the observations based on whether their predictions result in under predicted, over predicted or well predicted. This categorization is achieved by employing various quantiles across the distribution of prediction errors. Subsequently, both groups of wrongly predicted observations are analyzed separately to investigate the impact of each feature on each group. In essence, the study focuses on identifying variables that contribute to the observed biases within each group.\\

As previously discussed, the consideration of Shapley values within feature selection methods is not novel; however, their use at a more local level and their association with prediction errors represents a distinctive approach.\\

Together with this local use of Shapley values, the groups of well classified over predicted and under predicted observations are crucial for the algorithm. While we opted to employ quantiles as a means to establish these groups, it is important to note that this approach is not mandatory. There may exist other criteria that are equally suitable for creating this distinction among the observations. Alternative methods could be explored to effectively generate these groups, potentially providing additional insights for the analysis. \\

\subsection{The detailed algorithm}
The starting point of the algorithm is a given model and a set of training and validation data. All variables are considered and a backward selection strategy will be established, eliminating variables sequentially.\\

The first step is to train the model and obtain predictions on the validation set. Working individually with each of these predictions is not feasible, so three groups of observations are constructed based on the prediction error. For this purpose, two user-selected parameters are used, $q_{low},q_{high} \in [0,1]$, which correspond to two quantiles.

\begin{definition}\label{def:clasificacion}
Let \textbf{x} be the vector of explanatory variables of an observation $(\textbf{x}, y)$ of the validation set, err(\textbf{x},y) $= y-\hat y(\textbf{x})$ be the error of its prediction with the model considered, \textbf{err} be the vector of errors of all predictions, $Q_{low} = \text{Quantile}(\textbf{err}, q_{low})$ and $Q_{high}$ the analogue\footnote{Lower case is used for the quantile and upper case for the value associated with the quantile.}.
Let $q^\ast$ be the quantile such that $P(\textbf{err} \leq 0) = q^\ast$, $Q^\ast$ the value such that $\text{Quantile}(\textbf{err}, q^\ast) = Q^\ast$ (note that $Q^\ast$ is not necessarily 0).
The following are defined as $$ Q_{low}^\ast = 
\begin{cases}
Q_{low} \, &\text{if} \, 0 \in \left[Q_{low}, Q_{high} \right]\\
Q^\ast \, &\text{if} \, Q_{high} < 0\\
Q_{low} - (Q_{high} - Q^\ast) \ &\text{if} \, Q_{low} > 0
\end{cases}$$
$$
Q_{high}^\ast = 
\begin{cases}
Q_{high} \ &\text{if} \, 0 \in \left[Q_{low}, Q_{high} \right]\\
Q_{high} - (Q_{low} - Q^\ast)\ &\text{if} \, Q_{high} < 0\\
Q^\ast  \ &\text{if} \, Q_{low} > 0
\end{cases}
$$
It is said that:
\begin{itemize}
\item \textbf{x} is correctly predicted if err(\textbf{x},y) $\in \left[Q_{low}^\ast, Q_{high}^\ast \right]$
\item \textbf{x} is under predicted if err(\textbf{x},y) $\in \left(Q_{high}^\ast, +\infty\right)$
\item \textbf{x} is over predicted if err(\textbf{x},y) $\in \Big(-\infty, Q_{low}^\ast\Big)$
\end{itemize}
In this way, it is considered that $\left[Q_{low}^\ast, Q_{high}^\ast \right]$ is the correctly predicted space of errors while the complementary is wrongly predicted.
\end{definition}

\begin{figure}[h]
\centering
\includegraphics[scale=0.45]{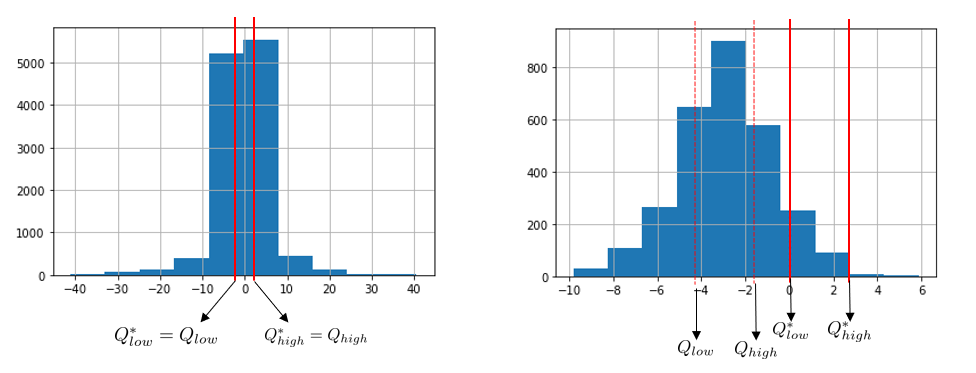}
\caption{The left-hand side shows that no translation is necessary since, given the quantiles chosen, the model does not show any significant bias. On the right, the model tends to overpredict, so this bias is penalised by shifting the quantiles to define the correctly predicted region.}
\label{fig:grupos_obs}
\end{figure}

Note that in the previous definition, in case that $0 \not \in \left[Q_{low}, Q_{high} \right]$, we are simply doing a translation of the quantiles in case the model is highly biased. Thus, the bias of the model is included in the poorly predicted space but keeping the same width between boundaries as in the original proposal. In Figure \ref{fig:grupos_obs} it is shown graphically over the distribution of the errors.\\

\begin{notation}\label{def:efecto}
Let $(\textbf{x}, y)$ be an observation from the validation set, ${\hat{y}}$ the model's prediction of that observation and SHAP$_{\text{var}}$ a function that returns the Shapley value of a variable for an observation. The effect of a variable, var, on an observation is denoted as $$\text{Effect}_{\text{var}, \textbf{x}} = \text{sgn}(\text{SHAP}_{\text{var}}(\textbf{x}, y, \hat{y})) \cdot \text{SHAP}_{\text{var}}(\textbf{x}, y, \hat{y})^2$$
\end{notation}

Note that the Shapley value could actually be taken directly, but taking the square increases the effect of the most influential characteristics.\\

The effect of a variable on a group (correctly predicted, over predicted, under predicted) can be derived as:

$$\text{Effect}_{\text{var}, \text{group}} \equiv \text{Ef}_{\text{var}, \text{group}} = \sum_{\textbf{x} \in \text{group}}\text{Effect}_{\text{var}, \textbf{x}}$$

\begin{definition}\label{def:influencia_negativa}
Let var be a variable, \textbf{err} the vector of errors of the predictions on the validation set, $q_2(\textbf{err})$ the median of the vector \textbf{err} and let C.P, O.P and U.P be the three groups described above, we define the negative influence of var, neg inf$_{\text{var}}$, as
$$\text{neg inf}_{\text{var}} = 
\begin{cases}
+\infty \ &\text{if } \, \displaystyle \sum_{group} \vert \text{Ef}_{\text{var}, \text{group}} \vert = 0\\\\

\vert \text{Ef}_{\text{var}, \text{O.P}} \vert - \left( \vert \text{Ef}_{\text{var}, \text{U.P}} \vert + \vert \text{Ef}_{\text{var}, \text{C.P}} \vert\right) \ &\text{if } \, q_2(\textbf{err}) < 0, \, \text{Ef}_{\text{var}, \text{O.P}} > 0, \\
&\text{Ef}_{\text{var}, \text{U.P}} > 0, \\
&\vert \text{Ef}_{\text{var}, \text{O.P}} \vert > \vert \text{Ef}_{\text{var}, \text{U.P}} \vert + \vert \text{Ef}_{\text{var}, \text{C.P}} \vert\\\\

\vert \text{Ef}_{\text{var}, \text{U.P}} \vert - \left( \vert \text{Ef}_{\text{var}, \text{O.P}} \vert + \vert \text{Ef}_{\text{var}, \text{C.P}} \vert\right) \ &\text{if } \, q_2(\textbf{err}) > 0, \, \text{Ef}_{\text{var}, \text{O.P}} > 0, \\
&\text{Ef}_{\text{var}, \text{U.P}} > 0, \\
&\vert \text{Ef}_{\text{var}, \text{U.P}} \vert > \vert \text{Ef}_{\text{var}, \text{O.P}} \vert + \vert \text{Ef}_{\text{var}, \text{C.P}} \vert\\\\

\vert \text{Ef}_{\text{var}, \text{U.P}} \vert + \vert \text{Ef}_{\text{var}, \text{O.P}} \vert - \vert \text{Ef}_{\text{var}, \text{C.P}} \vert \ &\text{if } \, \text{Ef}_{\text{var}, \text{O.P}} > 0, \, \text{Ef}_{\text{var}, \text{U.P}} < 0, \\
&\vert \text{Ef}_{\text{var}, \text{U.P}} \vert + \vert \text{Ef}_{\text{var}, \text{O.P}} \vert > \vert \text{Ef}_{\text{var}, \text{C.P}} \vert\\\\

0 &\text{otherwise}
\end{cases}
$$
\end{definition}

 The general idea of the previous definition is to study the effect that each variable has on each group of observations. Each of the cases encapsulates the following corresponding idea:
\begin{enumerate}
\item If a variable has no effect on the predictions, its negative influence is defined as infinite, as we are adding a variable to the model that does not contribute with any value and thus this variable is a candidate for overfitting.\\
\item In case the model is biased towards over predictions and the variable increases the value of over predictions (undesirable effect) and increases the value of under predictions (desirable effect), the negative influence is the difference of the absolute values of the undesirable minus the desirable effects. The correctly predicted group is always considered as a desired effect.\\
\item This is symmetrical to the previous case for the group of those under predicted.\\
\item In case the variable increases over predictions and decreases under predictions, the variable has an undesired effect for all but the correcty predicted ones.\\
\item In any other case, the variable has no negative influence.
\end{enumerate} 

Note the significance of segregating observations into different groups (C.P, O.P, and U.P). In each iteration, the model may be in a distinct bias position, consistently generating predictions that are persistently higher or lower. The model's position, as determined through $q_2(\textbf{err})$, influences whether a variable's effect can be negative or not. For instance, when $q_2(\textbf{err}) < 0$, indicating a general trend of overpredictions, a variable with a substantial effect on causing overpredictions (large $\vert \text{Ef}_{O.P}\vert $) becomes a candidate for having neg inf $> 0$. This indicates that the variable is one of the features contributing to the observed bias. On the contrary, when $q_2(\textbf{err}) > 0$, indicating a tendency of underpredictions, the same variable becomes a candidate to significantly aid the model, as $\vert \text{Ef}_{O.P} \vert$ in this case has a positive effect that helps increase the value of predictions, thereby mitigating the current bias, and its neg inf will be 0.\\

With these concepts defined, the algorithm can be fully described (Algorithm \ref{algorithm1})\\

\begin{algorithm}
\caption{New feature selection algorithm for regression problems}
\begin{algorithmic}[h]
\State \textbf{Input}: $(\textbf{X}, \textbf{y})_{\text{train}}, \, (\textbf{X}, \textbf{y})_{\text{val}}$, $model$, $n\_iter\_prev$, $q_{low}$, $q_{high}$, $metric$
\State \textbf{Output}: Set of selected features
\\  
\If{n\_iter\_prev $> 0$}
	\State Introduce a random variable to the training and validation sets\;
	\For{iteration in 1:$n\_iter\_prev$}
		\State Train a model with the new dataset\;
		\State Compute the global influence of each variable and store it for each iteration\;
	\EndFor
	\State Calculate the average of the global influences of the previous $n\_iter\_prev$ iterations\;
	\State Remove the random variable and the characteristics that have less overall influence than the random variable\;
\EndIf
\While{len(features selected) > 0 \textbf{and} len(features to remove) > 0}
	\State Train the model with the features selected up to now\;
	\State Compute Shapley values for all variables and predictions in the validation set\;
	\State Classify the observations of the validation set into correctly predicted, over predicted or under predicted\;
	\State Compute the effect of each variable on each observation of the validation set\;
	\State Compute the effect of each variable on each group of observations\;
	\State Establish the negative influence of each variable\;
	\State Compute the chosen \textit{metric} on the validation set\;
	\If{There is a variable with infinite negative influence}
		\State Delete all variables with infinite negative influence\;
	\Else
		\If{There is a variable with a non-zero negative influence}
			\State Delete the variable with the greatest negative influence\;
		\EndIf
	\EndIf
\EndWhile
\end{algorithmic}
\label{algorithm1}
\end{algorithm}

The algorithm is divided into two phases, a preliminary and optional phase and the main component. In the first preprocessing phase, a random variable is introduced into the dataset, specifically, a permutation of the most influential variable according to the mean of the absolute values of the Shapley values between the original variables. Once included in the dataset, the global influence of each variable is recalculated a certain number of times specified by the user, where in each iteration the random seed that determines the learning process of the algorithm is modified to obtain different influences. At the end of the process, all those variables that have a global influence less than or equal to that of the new variable introduced are removed. Including this first phase seems to improve the results when the number of variables eliminated is not too large, since it deletes variables that only seem to introduce noise. In case a large number of variables are excluded in this phase, it is recommended not to apply it, since the importance or role of each variable in the decisions of the model may not be entirely clear.\\ 

The second component corresponds to the core part of the algorithm, where the effect of each variable is analyzed when making predictions. This is a backward feature selection scheme, i.e., variables are sequentially removed until a criterion is no longer met. To eliminate a variable, first the predictions in the validation set are classified according to the error made and the $q_{low}$ and $q_{high}$ quantiles chosen (Definition \ref{def:clasificacion}). Then, according to the effect of each variable on each group (Notation \ref{def:efecto}), the negative influence is computed (Definition \ref{def:influencia_negativa}). If there are variables that have no effect on the predictions, that is, with infinite negative influence, in this iteration all those that meet this property are eliminated. If all the variables have some effect on the prediction, the variable with the greatest negative influence is eliminated. The process ends when there is no variable with non-zero negative influence or when there are no more features left. At the end of each iteration a metric (MAE, MSE, R$^2$, etc.) can be computed on the predictions of the validation set in an informative way, although it is not used to make decisions during the process. The feature set that obtained the best value of the metric is returned. Supposing that a value very similar to the best could be obtained with a notably smaller number of variables, which would also constitute a reasonable final feature set, the user could decide to consider it.\\

\subsection{Remarks on the algorithm}

As previously mentioned, the shift is not detected explicitly, but in the presence of a change with a negative influence on a variable, the proposed algorithm is able to detect its undesired effect and evaluate whether the preservation of the characteristic in the model is really beneficial.\\

One of the challenges in identifying variables that can cause behavior changes arises from the presence of correlated variables that collectively contribute to the occurrence of the change. Our algorithm does not directly consider this issue. Addressing it is not straightforward, as it would entail handling both groups of features and individualized variables, which may not be feasible when working automatically without expert knowledge. 
However, the fact that we are considering as the final set of features the one with the best metric on the validation set can indirectly tackle the problem. 
In the case of several correlated variables that are individually worsen the model but collectively improving it, one could detect such situations analyzing the increases in the MAE variations between iterations in the validation set (as depicted in Figure \ref{fig:variacion_mae}). In the opposite situation, in which the correlated variables are individually improving the model but collectively worsen it, it may happen that these variables are kept as we are not treating the possible influence of the different groups of variables. Employing a cross-validation strategy, as commonly practiced in Machine Learning tasks, may help detect this issue and facilitate the selection of a set of variables that are expected to perform well on the test set.\\

Regarding the computational complexity of the algorithm, four factors can influence it: the complexity associated with model training, the complexity of evaluating new observations, the efficiency in estimating the Shapley value, and the number of considered features. While these factors pose no issues for algorithms based on decision trees, linear regression models, or generalized linear models, they may make the algorithm impractical for neural networks. To analize this case, consider, for example, the algorithm of \cite{ancona2019explaining}. It gives an approximation of the Shapley values in $\mathcal{O}(M^2)$ network evaluations, where M is the number of features, which is usually dominated by the computational complexity associated to the network training. When dealing with complex tasks requiring networks with numerous parameters, conducting all the mentioned phases in each iteration of the algorithm can become infeasible, despite each step being computationally viable in polynomial time. Therefore, this limitation should be taken into consideration.\\

Let us stress why this algorithm is designed to be robust to concept shift situations. The variable importance is not the only metric that is taken into account. In fact, the possible bias generated by a variable is the main focus of the algorithm. If a variable is constantly generating a bias in the model predictions, which is common in concept shift contexts, then that variable is actually discarded from the possible explanatory variables, even if it has a great impact on the model. It is crucial to emphasize that, similar to constant model retraining in real-world applications to adapt to changing conditions, the reintegration of, at least, these important features (based on a global metric like the mean of the absolute Shapley values across all observations) should be assessed. If the feature genuinely influences the target variable, even in the presence of concept shifts affecting that explanatory variable, there is a possibility that, with a substantial amount of new collected data related to the shift, it could once again have a positive effect on the predictions and, therefore, in model performance. In such instances, reintroducing it into the model feature set should be considered.\\

\section{Experiments}\label{sec:experimentos}

To analyze the effectiveness of the algorithm, the proposed method was compared with other feature selection algorithms using different configurations. Specifically, the quantile configurations $(0.25, 0.75)$, $(0.2, 0.8)$, $(0.15, 0.85)$, $(0.1, 0.9)$, $(0.05, 0.95)$, which correspond to considering from 50\% of poorly predicted observations in the development of the algorithm to 10\%, were analyzed. In addition, 30 iterations of the preprocessing phase were applied and the MAE is used to select the final set of features.\\

The other algorithms evaluated are considered as references in the variable selection methods: Boruta, PIMP and Lasso as traditional methods and Boruta-Shap, Shapicant and Powershap based on Shapley values. In case they use validation and training sets to make the variable selection, the same are introduced in all cases, if only one dataset is used, validation and training are introduced as one. In the particular case of the Lasso regularization, the variables are previously normalized by the maximum and minimum, and four different values of the regularization constant are considered: $0.01; 0.001; 0.0001; 0.00001$.\\

Following \cite{verhaeghe2022powershap}, a CatBoost estimator  \citep{NEURIPS2018_14491b75} with 250 iterations is applied on all datasets and on all variable selection algorithms, except Lasso regression to select features. On the test set, we analyze the MAE, RMSE and R$^2$ of 50 different runs where the seed is varied to obtain different results and to observe the stability of the results with the selected variables. In contrast to the Verhaeghe's methodology, the 50 seeds are fixed in advance and the order of the variables is always entered alphabetically, otherwise different results could be obtained with the same set of features\footnote{To ensure the veracity and reproducibility of the results, the different databases, the algorithm code and the results can be found at \url{https://github.com/CCaribe9/SHAPEffects}}.\\

Two different situations are analyzed: selection of variables with the presence of concept shift and selection of variables in standard situations.\\

\subsection{Concept shift}

\subsubsection{Synthetic experiments}

\begin{figure}[H]
\centering
\includegraphics[scale=0.4]{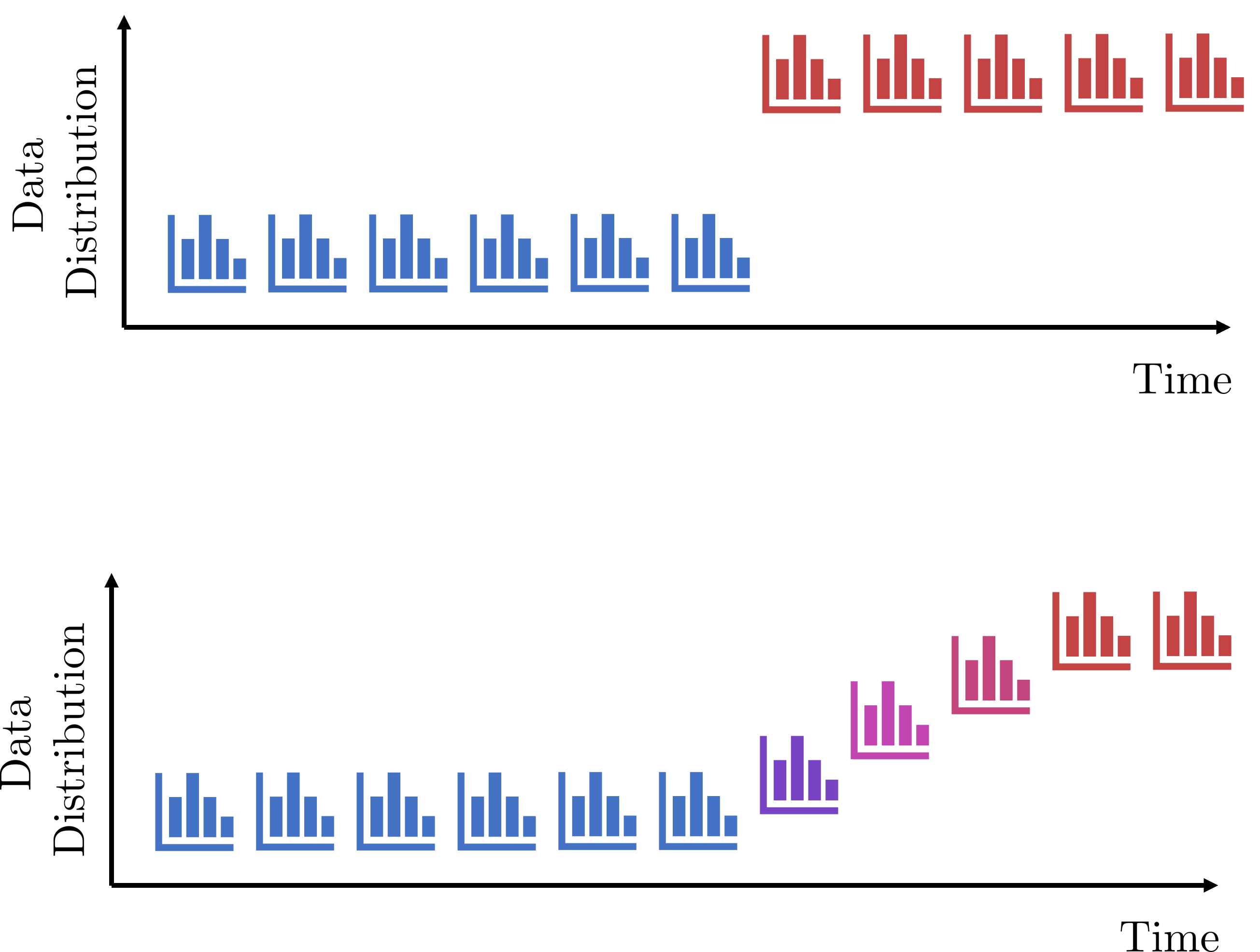}
\caption{\textbf{Above} is a sudden shift situation. \textbf{Below} is an incremental shift situation.}
\label{fig:drift_types}
\end{figure}

We believe that analizying the method in fully controlled toy datasets is necessary to test the effectiveness of the method. To this end, two of the most typical dataset shift situations have been recreated: a sudden shift and an incremental shift \citep{lu2018learning}, represented in Figure \ref{fig:drift_types}.

We will try to fit the following objective function:
\begin{center}
\begin{align*}
f(\textbf{x}_t) &= 2x_{1, t} + \lambda_1 x_{2, t}^2 + 3 \sin (2\pi x_{3, t}) - 0.4 x_{4, t} + \lambda_2 x_{5, t}^2\\
&+2x_{1, t-1} + \lambda_1 x_{2, t-1}^2 + 3 \sin (2\pi x_{3, t-1}) - 0.4 x_{4, t-1} + \lambda_2 x_{5, t-1}^2\\&+ \varepsilon_t
\end{align*}
\end{center}

where $\textbf{x}_t = (x_{1, t}, \dots, x_{5, t}, \dots, x_{10, t}) \in \mathbb{R}^{10}, \lambda_1, \lambda_2 \in \mathbb{R}$. Note that only the first five variables are informative. The scalars $\lambda_1$ and $\lambda_2$  are employed to generate diverse shift scenarios, while $\varepsilon_t$ represents noise. This function has been chosen in a way that linearities and nonlinearites are both included. For nonlinearities standard simple functions has been selected. An autoregressive component has been considered, aligning with typical time series problems. Different coefficients have chosen to visualize how the algorithm performs with various scales of the features.\\

We generate 30000 samples of $x_i \sim U(0, 1), \, i = 1, \dots, 10$. We model the noise as a $N(0, 0.01)$. We also include as a possible explanatory variable the first lag of the objective variable.\\

Let $\lambda_1^a \in \left\lbrace -10, -1, -0.1 \right\rbrace , \, \lambda_1^b \in \left\lbrace -4, -0.4, -0.04 \right\rbrace , \, \lambda_2^a \in \left\lbrace 10, 1, 0.1 \right\rbrace , \, \lambda_2^b \in \left\lbrace -25, -2.5, -0.25 \right\rbrace$. In order to recreate a sudden concept shift, we define\\
\begin{center}
$
\lambda_1 =
\begin{cases}
\lambda_1^a, &\text{for the first 20000 samples}\\
\lambda_1^b, &\text{for the last 10000 samples}\\
\end{cases}
$\\[10pt]
$
\lambda_2 =
\begin{cases}
\lambda_2^a, &\text{for the first 20000 samples}\\
\lambda_2^b, &\text{for the last 10000 samples}\\
\end{cases}
$
\end{center}
\vskip 5pt
To recreate an incremental concept shift situation, if we call index to the sample number, we define:\\
\begin{center}
$
\lambda_1 =
\begin{cases}
\lambda_1^a, &\text{for the first 20000 samples}\\
\dfrac{(\lambda_1^b - \lambda_1^a)(\text{index} - 20000) + 10000\lambda_1^a}{10000}, &\text{if index } \in (20000, 25000)\\
\lambda_1^b, &\text{for the last 5000 samples}\\
\end{cases}
$\\[10pt]
$
\lambda_2 =
\begin{cases}
\lambda_2^a, &\text{for the first 20000 samples}\\
\dfrac{(\lambda_2^b - \lambda_2^a)(\text{index} - 20000) + 10000\lambda_2^a}{10000}, &\text{if index }  \in (20000, 25000)\\
\lambda_2^b, &\text{for the last 5000 samples}\\
\end{cases}
$
\end{center}
\vskip 5pt
Every combination of $\lambda_1^a, \lambda_1^b, \lambda_2^a, \lambda_2^b$  is taken into account, resulting in a total of 81 potential scenarios for both types of shifts. This aspect holds particular interest as it enables the analysis of the system's behavior when the impacted variables exhibit varying degrees of significance.\\

For both cases, the first 20000 samples are used for training, the next 5000 for validation and the last 5000 as test. The different sets are considered following a temporal order, as in time series problems. The shift occurs inside the validation set, which is the first moment in which the algorithms could detect the change of behaviour. The  sudden shift is considered at the start of the validation set, so there is bigger contrast between the sudden shift and the incremental shift situation, which happens throughout the entire validation set. We compute the difference between the mean MAE/RMSE/R$^2$ of the proposed algorithm with every other method, although the standard deviation, the minimum value and the maximum value are also measured. 

\begin{figure}[h]
\centering
\includegraphics[scale=0.3]{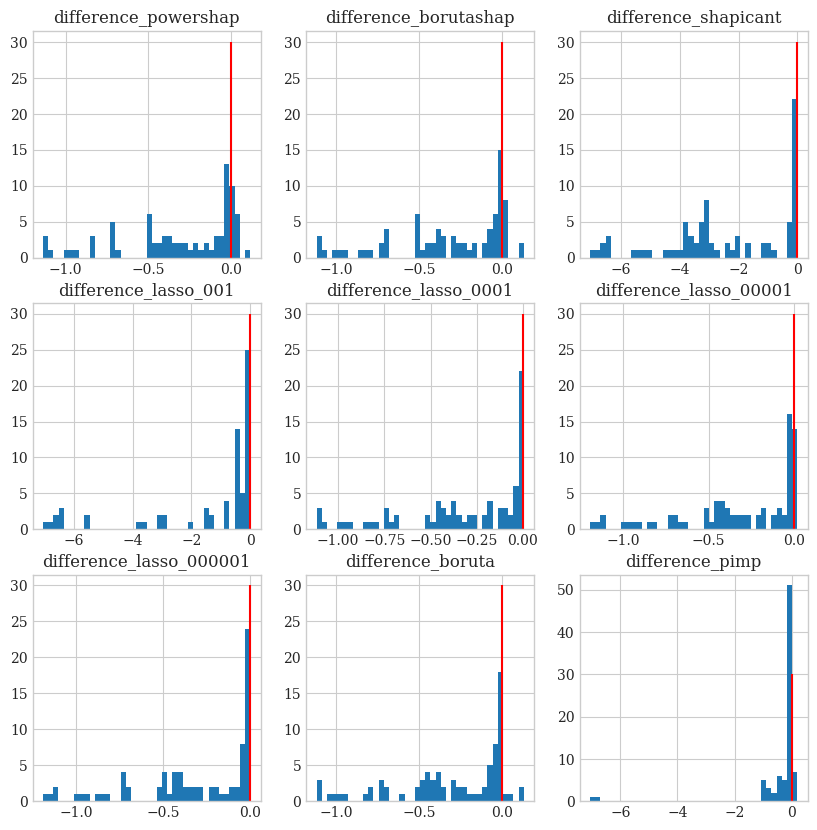}
\caption{Histograms (for the 81 cases) of the difference of the mean MAE of the proposed method with every other algorithm for the sudden shift case}
\label{fig:results_sudden}
\end{figure}

\begin{figure}[h]
\centering
\includegraphics[scale=0.3]{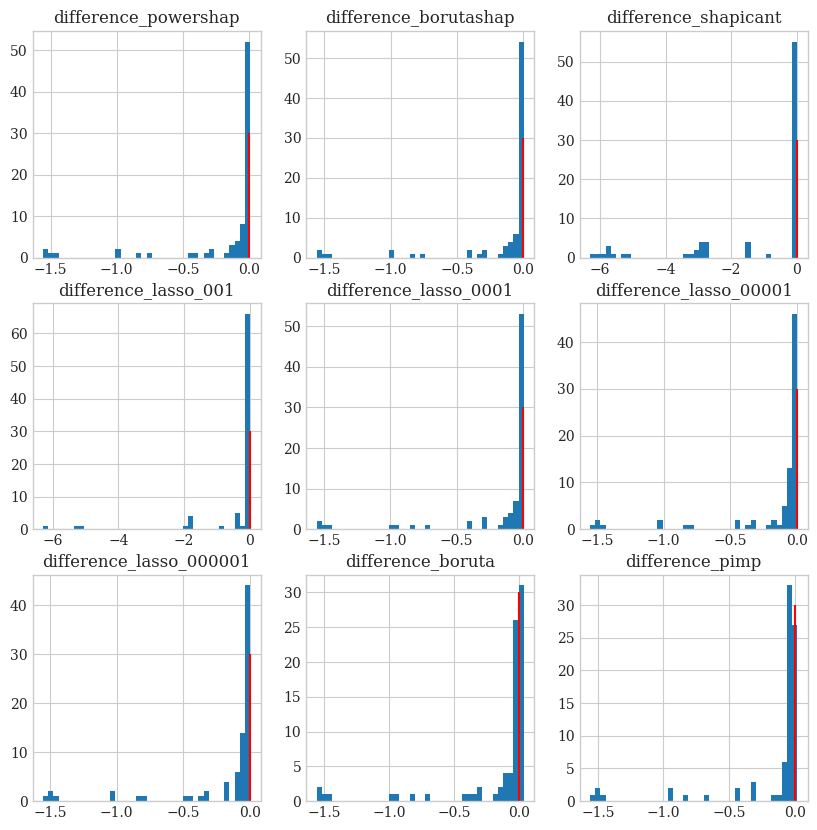}
\caption{Histograms (for the 81 cases) of the difference of the mean MAE of the proposed method with every other algorithm for the incremental shift case}
\label{fig:results_incremental}
\end{figure}

\paragraph*{Results}The proposed method is called SHAPEffects. The results are graphically described by the histogram of each one of these differences in Figures \ref{fig:results_sudden} and \ref{fig:results_incremental}.\\

Only MAE outcomes are presented, as the results for RMSE and R$^2$ are equivalent. The obtained results are highly encouraging, given that the difference is predominantly negative (smaller MAE). In fact, when that difference is positive, it is practically negligible. These instances arise when the importance of one or both variables subjected to the shift is almost insignificant, indicated by comparable small coefficient values. The analysis reveals that the disparity between the two types of shifts is minimal. Moreover, the proposed algorithm consistently achieves superior results compared to other methods. The marginal discrepancies observed between the two shift types arise when the influence of the variables is relatively limited. In the case of incremental shifts, where a gradual change occurs, the proposed algorithm may not find beneficial to eliminate some of the variables that are undergoing the shift in the validation data, as at the beginning of the shift the variables may be contributing correctly to the outcomes. This behavior aligns with the approach followed by the other algorithms, contributing to the small differences observed (see Tables \ref{cuadro:sintetico5} and \ref{cuadro:sintetico6}).\\

\begin{figure}[h]
\centering
\includegraphics[scale=0.4]{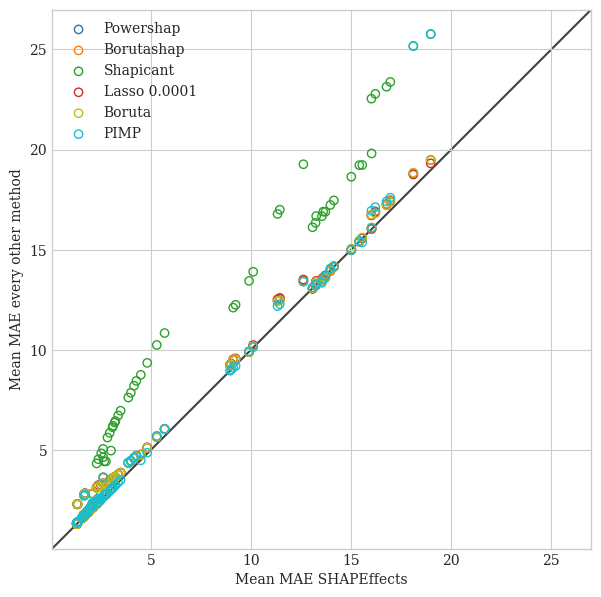}
\caption{Mean MAE of the best SHAPEffects configuration vs mean MAE of every other method for the sudden shift case}
\label{fig:results2_sudden}
\end{figure}

\begin{figure}[h]
\centering
\includegraphics[scale=0.4]{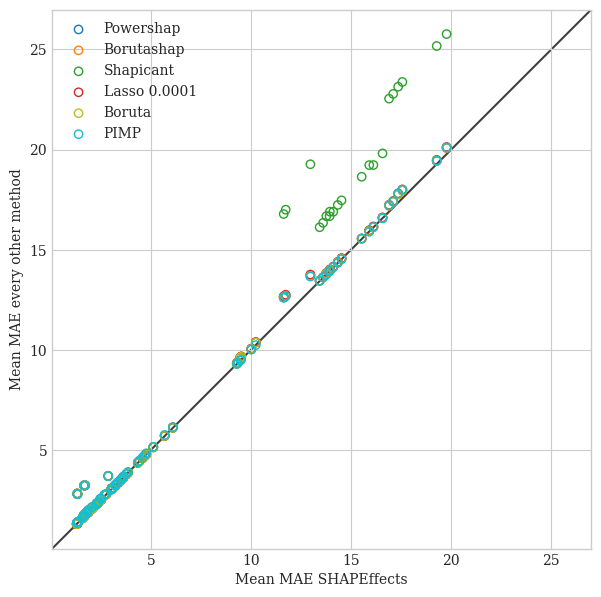}
\caption{Mean MAE of the best SHAPEffects configuration vs mean MAE of every other method for the incremental shift case}
\label{fig:results2_incremental}
\end{figure}

For a more individualized evaluation rather than a collective comparison, the mean MAE of the optimal SHAPEffects configuration is compared with the mean MAE of each alternative method, and subsequently visualized on a scatter plot (Figures \ref{fig:results2_sudden} and \ref{fig:results2_incremental}). Similarly, in the case of the Lasso, the best configuration is selected, which from the previous figures is $\lambda = 0.0001$.\\

It can be seen that when the MAE is low (when the coefficients of the changing variables are very small) all methods exhibit similar performance, as previously stated out. However, as the error increases, which corresponds to a greater increase in the influence of the variables that undergo the change in behaviour, there are situations in which the MAE is notably greater than that obtained by our proposal for all the algorithms, which can be seen when we see that no method exceeds the line $y=x$. These behaviours are more evident in the case of the sudden shift, although they are also visible in the incremental case.\\

It is important to note that the cases in which the proposed algorithm performs notably better than the other algorithms (when the points are separated from the diagonal), is when one of the two variables undergoing the change is removed. Specifically, when $x_5$ is eliminated, which is the most relevant one. The rest of the algorithms do not drop these variables when they take on a notable importance (Tables \ref{cuadro:sintetico1}, \ref{cuadro:sintetico2}, \ref{cuadro:sintetico3} and \ref{cuadro:sintetico4}).\\

Furthermore, we look in detail at what we believe to be the three most representative cases:
\begin{enumerate}
    \item $\lambda_1^a = -10, \lambda_1^b = -4, \lambda_2^a = 10, \lambda_2^b = -25$ (Table \ref{cuadro:sintetico1} and \ref{cuadro:sintetico2})
    \item $\lambda_1^a = -1, \lambda_1^b = -0.4, \lambda_2^a = 1, \lambda_2^b = -2.5$ (Table \ref{cuadro:sintetico3} and \ref{cuadro:sintetico4})
    \item $\lambda_1^a = -0.1, \lambda_1^b = -0.04, \lambda_2^a = 0.1, \lambda_2^b = -0.25$ (Table \ref{cuadro:sintetico5} and \ref{cuadro:sintetico6})
\end{enumerate}

\begin{table}[h]
\resizebox{\textwidth}{!}{
\begin{tabular}{c|llll|llll|llll}
\multirow{2}{*}{\textbf{Algorithm}}                              & \multicolumn{4}{c|}{\textbf{MAE}}                                                                                                           & \multicolumn{4}{c|}{\textbf{RMSE}}                                                                                                          & \multicolumn{4}{c}{\textbf{R}}                                                                                                             \\
                                                                 & \multicolumn{1}{c}{\textbf{Mean}} & \multicolumn{1}{c}{\textbf{Std}} & \multicolumn{1}{c}{\textbf{Max}} & \multicolumn{1}{c|}{\textbf{Min}} & \multicolumn{1}{c}{\textbf{Mean}} & \multicolumn{1}{c}{\textbf{Std}} & \multicolumn{1}{c}{\textbf{Max}} & \multicolumn{1}{c|}{\textbf{Min}} & \multicolumn{1}{c}{\textbf{Mean}} & \multicolumn{1}{c}{\textbf{Std}} & \multicolumn{1}{c}{\textbf{Max}} & \multicolumn{1}{c}{\textbf{Min}} \\ \hline
Powershap                                                        & 13.44                             & 3.99e-02                         & 13.54                            & 13.31                             & 17.18                             & 4.90e-02                         & 17.30                            & 17.05                             & -1.41                             & 1.38e-02                         & -1.38                            & -1.45                            \\
Boruta-Shap                                                      & 13.41                             & 2.70e-02                         & 13.47                            & 13.36                             & 17.14                             & 3.95e-02                         & 17.24                            & 17.06                             & -1.40                             & 1.11e-02                         & -1.38                            & -1.43                            \\
Shapicant                                                        & 19.27                             & 3.94e-03                         & 19.27                            & 19.26                             & 22.30                             & 1.02e-02                         & 22.32                            & 22.27                             & -3.07                             & 3.73e-03                         & -3.06                            & -3.07                            \\
Boruta                                                           & 13.41                             & 2.70e-02                         & 13.47                            & 13.36                             & 17.14                             & 3.95e-02                         & 17.24                            & 17.06                             & -1.40                             & 1.11e-02                         & -1.38                            & -1.43                            \\
PIMP                                                             & 13.39                             & 2.84e-02                         & 13.46                            & 13.33                             & 17.11                             & 3.64e-02                         & 17.21                            & 17.02                             & -1.39                             & 1.02e-02                         & -1.37                            & -1.42                            \\
\begin{tabular}[c]{@{}c@{}}Best Lasso \\ (0.001)\end{tabular}    & 13.41                             & 2.70e-02                         & 13.47                            & 13.36                             & 17.14                             & 0.04                             & 17.24                            & 17.06                             & -1.40                             & 1.11e-02                         & -1.38                            & -1.43                            \\
\begin{tabular}[c]{@{}c@{}}SHAPEffects\\ (0.25-075)\end{tabular} & 12.89                             & 4.64e-02                         & 12.97                            & 12.78                             & 15.89                             & 5.14e-02                         & 15.98                            & 15.76                             & -1.06                             & 1.33e-02                         & -1.03                            & -1.09                            \\
\begin{tabular}[c]{@{}c@{}}SHAPEffects\\ (0.2-08)\end{tabular}   & 12.73                             & 4.00e-02                         & 12.80                            & 12.63                             & 15.73                             & 4.41e-02                         & 15.81                            & 15.62                             & -1.02                             & 1.13e-02                         & -0.99                            & -1.04                            \\
\begin{tabular}[c]{@{}c@{}}SHAPEffects\\ (0.15-085)\end{tabular} & 12.73                             & 4.00e-02                         & 12.80                            & 12.63                             & 15.73                             & 4.41e-02                         & 15.81                            & 15.62                             & -1.02                             & 1.13e-02                         & -0.99                            & -1.04                            \\
\begin{tabular}[c]{@{}c@{}}SHAPEffects\\ (0.1-0.9)\end{tabular}  & \textbf{12.61}                    & 2.90e-02                         & \textbf{12.68}                   & \textbf{12.52}                    & \textbf{15.60}                    & 3.30e-02                         & \textbf{15.67}                   & \textbf{15.49}                    & \textbf{-0.99}                    & 8.41e-03                         & \textbf{-0.96}                   & \textbf{-1.01}                   \\
\begin{tabular}[c]{@{}c@{}}SHAPEffects\\ (0.05-095)\end{tabular} & 12.80                             & 4.63e-02                         & 12.91                            & 12.69                             & 15.82                             & 5.07e-02                         & 15.94                            & 15.69                             & -1.05                             & 1.31e-02                         & -1.01                            & -1.08                           
\end{tabular}}
\caption{Test results on the first case for the sudden shift context}
\label{cuadro:sintetico1}
\end{table}

\begin{table}[h]
\resizebox{\textwidth}{!}{
\begin{tabular}{c|llll|llll|llll}
\multirow{2}{*}{\textbf{Algorithm}}                              & \multicolumn{4}{c|}{\textbf{MAE}}                                                                                                           & \multicolumn{4}{c|}{\textbf{RMSE}}                                                                                                          & \multicolumn{4}{c}{\textbf{R}}                                                                                                             \\
                                                                 & \multicolumn{1}{c}{\textbf{Mean}} & \multicolumn{1}{c}{\textbf{Std}} & \multicolumn{1}{c}{\textbf{Max}} & \multicolumn{1}{c|}{\textbf{Min}} & \multicolumn{1}{c}{\textbf{Mean}} & \multicolumn{1}{c}{\textbf{Std}} & \multicolumn{1}{c}{\textbf{Max}} & \multicolumn{1}{c|}{\textbf{Min}} & \multicolumn{1}{c}{\textbf{Mean}} & \multicolumn{1}{c}{\textbf{Std}} & \multicolumn{1}{c}{\textbf{Max}} & \multicolumn{1}{c}{\textbf{Min}} \\ \hline
Powershap                                                        & 13.71                             & 1.96e-02                         & 13.75                            & 13.66                             & 17.52                             & 0.03                             & 17.58                            & 17.48                             & -1.51                             & 7.25e-03                         & -1.50                            & -1.53                            \\
Boruta-Shap                                                      & 13.71                             & 1.96e-02                         & 13.75                            & 13.66                             & 17.52                             & 0.03                             & 17.58                            & 17.48                             & -1.51                             & 7.25e-03                         & -1.50                            & -1.53                            \\
Shapicant                                                        & 19.27                             & 4.07e-03                         & 19.27                            & 19.26                             & 22.30                             & 0.01                             & 22.32                            & 22.27                             & -3.07                             & 3.83e-03                         & -3.06                            & -3.07                            \\
Boruta                                                           & 13.68                             & 2.07e-02                         & 13.71                            & 13.61                             & 17.48                             & 0.03                             & 17.53                            & 17.38                             & -1.50                             & 7.83e-03                         & -1.47                            & -1.51                            \\
PIMP                                                             & 13.65                             & 1.84e-02                         & 13.69                            & 13.61                             & 17.44                             & 0.03                             & 17.49                            & 17.38                             & -1.49                             & 7.24e-03                         & -1.47                            & -1.50                            \\
\begin{tabular}[c]{@{}c@{}}Best Lasso \\ (0.001)\end{tabular}    & 13.68                             & 2.07e-02                         & 13.71                            & 13.61                             & 17.48                             & 0.03                             & 17.53                            & 17.38                             & -1.50                             & 7.83e-03                         & -1.47                            & -1.51                            \\
\begin{tabular}[c]{@{}c@{}}SHAPEffects\\ (0.25-075)\end{tabular} & \textbf{12.97}                    & 2.54e-02                         & \textbf{13.04}                   & \textbf{12.91}                    & \textbf{15.98}                    & 0.03                             & \textbf{16.06}                   & \textbf{15.92}                    & \textbf{-1.09}                    & 7.43e-03                         & \textbf{-1.07}                   & \textbf{-1.11}                   \\
\begin{tabular}[c]{@{}c@{}}SHAPEffects\\ (0.2-08)\end{tabular}   & \textbf{12.97}                    & 2.54e-02                         & \textbf{13.04}                   & \textbf{12.91}                    & \textbf{15.98}                    & 0.03                             & \textbf{16.06}                   & \textbf{15.92}                    & \textbf{-1.09}                    & 7.43e-03                         & \textbf{-1.07}                   & \textbf{-1.11}                   \\
\begin{tabular}[c]{@{}c@{}}SHAPEffects\\ (0.15-085)\end{tabular} & \textbf{12.97}                    & 2.54e-02                         & \textbf{13.04}                   & \textbf{12.91}                    & \textbf{15.98}                    & 0.03                             & \textbf{16.06}                   & \textbf{15.92}                    & \textbf{-1.09}                    & 7.43e-03                         & \textbf{-1.07}                   & \textbf{-1.11}                   \\
\begin{tabular}[c]{@{}c@{}}SHAPEffects\\ (0.1-0.9)\end{tabular}  & \textbf{12.97}                    & 2.54e-02                         & \textbf{13.04}                   & \textbf{12.91}                    & \textbf{15.98}                    & 0.03                             & \textbf{16.06}                   & \textbf{15.92}                    & \textbf{-1.09}                    & 7.43e-03                         & \textbf{-1.07}                   & \textbf{-1.11}                   \\
\begin{tabular}[c]{@{}c@{}}SHAPEffects\\ (0.05-095)\end{tabular} & \textbf{12.97}                    & 2.54e-02                         & \textbf{13.04}                   & \textbf{12.91}                    & \textbf{15.98}                    & 0.03                             & \textbf{16.06}                   & \textbf{15.92}                    & \textbf{-1.09}                    & 7.43e-03                         & \textbf{-1.07}                   & \textbf{-1.11}                  
\end{tabular}}
\caption{Test results on the first case for the incremental shift context}
\label{cuadro:sintetico2}
\end{table}

\begin{table}[h]
\resizebox{\textwidth}{!}{
\begin{tabular}{c|llll|llll|llll}
\multirow{2}{*}{\textbf{Algorithm}}                              & \multicolumn{4}{c|}{\textbf{MAE}}                                                                                                           & \multicolumn{4}{c|}{\textbf{RMSE}}                                                                                                          & \multicolumn{4}{c}{\textbf{R}}                                                                                                             \\
                                                                 & \multicolumn{1}{c}{\textbf{Mean}} & \multicolumn{1}{c}{\textbf{Std}} & \multicolumn{1}{c}{\textbf{Max}} & \multicolumn{1}{c|}{\textbf{Min}} & \multicolumn{1}{c}{\textbf{Mean}} & \multicolumn{1}{c}{\textbf{Std}} & \multicolumn{1}{c}{\textbf{Max}} & \multicolumn{1}{c|}{\textbf{Min}} & \multicolumn{1}{c}{\textbf{Mean}} & \multicolumn{1}{c}{\textbf{Std}} & \multicolumn{1}{c}{\textbf{Max}} & \multicolumn{1}{c}{\textbf{Min}} \\ \hline
Powershap                                                        & 1.79                              & 2.63e-03                         & 1.79                             & 1.78                              & 2.23                              & 2.79e-03                         & 2.23                             & 2.22                              & 0.53                              & 1.18e-03                         & 0.53                             & 0.53                             \\
Boruta-Shap                                                      & 1.79                              & 2.63e-03                         & 1.79                             & 1.78                              & 2.23                              & 2.79e-03                         & 2.23                             & 2.22                              & 0.53                              & 1.18e-03                         & 0.53                             & 0.53                             \\
Shapicant                                                        & 1.79                              & 2.43e-03                         & 1.80                             & 1.78                              & 2.23                              & 2.33e-03                         & 2.24                             & 2.22                              & 0.53                              & 9.90e-04                         & 0.53                             & 0.52                             \\
Boruta                                                           & 1.79                              & 2.43e-03                         & 1.80                             & 1.78                              & 2.23                              & 2.33e-03                         & 2.24                             & 2.22                              & 0.53                              & 9.90e-04                         & 0.53                             & 0.52                             \\
PIMP                                                             & 1.74                              & 1.55e-03                         & 1.74                             & 1.73                              & 2.16                              & 1.60e-03                         & 2.16                             & 2.15                              & 0.56                              & 6.59e-04                         & 0.56                             & 0.55                             \\
\begin{tabular}[c]{@{}c@{}}Best Lasso \\ (0.001)\end{tabular}    & 1.79                              & 2.63e-03                         & 1.79                             & 1.78                              & 2.23                              & 2.79e-03                         & 2.23                             & 2.22                              & 0.53                              & 1.18e-03                         & 0.53                             & 0.53                             \\
\begin{tabular}[c]{@{}c@{}}SHAPEffects\\ (0.25-075)\end{tabular} & \textbf{1.70}                     & 2.60e-03                         & \textbf{1.71}                    & \textbf{1.69}                     & \textbf{2.11}                     & 2.70e-03                         & \textbf{2.11}                    & \textbf{2.10}                     & \textbf{0.58}                     & 1.08e-03                         & \textbf{0.58}                    & \textbf{0.58}                    \\
\begin{tabular}[c]{@{}c@{}}SHAPEffects\\ (0.2-08)\end{tabular}   & \textbf{1.70}                     & 2.25e-03                         & \textbf{1.70}                    & \textbf{1.69}                     & \textbf{2.10}                     & 2.65e-03                         & \textbf{2.11}                    & \textbf{2.10}                     & \textbf{0.58}                     & 1.06e-03                         & \textbf{0.58}                    & \textbf{0.58}                    \\
\begin{tabular}[c]{@{}c@{}}SHAPEffects\\ (0.15-085)\end{tabular} & \textbf{1.70}                     & 2.56e-03                         & \textbf{1.70}                    & \textbf{1.69}                     & \textbf{2.10}                     & 2.76e-03                         & \textbf{2.11}                    & \textbf{2.09}                     & \textbf{0.58}                     & 1.10e-03                         & \textbf{0.58}                    & \textbf{0.58}                    \\
\begin{tabular}[c]{@{}c@{}}SHAPEffects\\ (0.1-0.9)\end{tabular}  & \textbf{1.70}                     & 2.83e-03                         & \textbf{1.70}                    & \textbf{1.69}                     & \textbf{2.11}                     & 3.08e-03                         & \textbf{2.11}                    & \textbf{2.10}                     & \textbf{0.58}                     & 1.23e-03                         & \textbf{0.58}                    & \textbf{0.58}                    \\
\begin{tabular}[c]{@{}c@{}}SHAPEffects\\ (0.05-095)\end{tabular} & \textbf{1.70}                     & 2.56e-03                         & \textbf{1.70}                    & \textbf{1.69}                     & \textbf{2.10}                     & 2.76e-03                         & \textbf{2.11}                    & \textbf{2.09}                     & \textbf{0.58}                     & 1.10e-03                         & \textbf{0.58}                    & \textbf{0.58}                   
\end{tabular}}
\caption{Test results on the second case for the sudden shift context}
\label{cuadro:sintetico3}
\end{table}

\begin{table}[h]
\resizebox{\textwidth}{!}{
\begin{tabular}{c|llll|llll|llll}
\multirow{2}{*}{\textbf{Algorithm}}                              & \multicolumn{4}{c|}{\textbf{MAE}}                                                                                                           & \multicolumn{4}{c|}{\textbf{RMSE}}                                                                                                          & \multicolumn{4}{c}{\textbf{R}}                                                                                                             \\
                                                                 & \multicolumn{1}{c}{\textbf{Mean}} & \multicolumn{1}{c}{\textbf{Std}} & \multicolumn{1}{c}{\textbf{Max}} & \multicolumn{1}{c|}{\textbf{Min}} & \multicolumn{1}{c}{\textbf{Mean}} & \multicolumn{1}{c}{\textbf{Std}} & \multicolumn{1}{c}{\textbf{Max}} & \multicolumn{1}{c|}{\textbf{Min}} & \multicolumn{1}{c}{\textbf{Mean}} & \multicolumn{1}{c}{\textbf{Std}} & \multicolumn{1}{c}{\textbf{Max}} & \multicolumn{1}{c}{\textbf{Min}} \\ \hline
Powershap                                                        & 1.79                              & 4.44e-03                         & 1.80                             & 1.78                              & 2.23                              & 3.76e-03                         & 2.24                             & 2.22                              & 0.53                              & 1.60e-03                         & 0.53                             & 0.52                             \\
Boruta-Shap                                                      & 1.79                              & 4.44e-03                         & 1.80                             & 1.78                              & 2.23                              & 3.76e-03                         & 2.24                             & 2.22                              & 0.53                              & 1.60e-03                         & 0.53                             & 0.52                             \\
Shapicant                                                        & 1.79                              & 4.10e-03                         & 1.80                             & 1.78                              & 2.23                              & 3.04e-03                         & 2.24                             & 2.23                              & 0.53                              & 1.29e-03                         & 0.53                             & 0.52                             \\
Boruta                                                           & 1.79                              & 4.10e-03                         & 1.80                             & 1.78                              & 2.23                              & 3.04e-03                         & 2.24                             & 2.23                              & 0.53                              & 1.29e-03                         & 0.53                             & 0.52                             \\
PIMP                                                             & 1.74                              & 1.49e-03                         & 1.74                             & 1.73                              & 2.16                              & 1.39e-03                         & 2.16                             & 2.15                              & 0.56                              & 5.71e-04                         & 0.56                             & 0.56                             \\
\begin{tabular}[c]{@{}c@{}}Best Lasso \\ (0.001)\end{tabular}    & 1.79                              & 4.44e-03                         & 1.80                             & 1.78                              & 2.23                              & 3.76e-03                         & 2.24                             & 2.22                              & 0.53                              & 1.60e-03                         & 0.53                             & 0.52                             \\
\begin{tabular}[c]{@{}c@{}}SHAPEffects\\ (0.25-075)\end{tabular} & \textbf{1.70}                     & 2.92e-03                         & 1.71                             & \textbf{1.69}                     & 2.11                              & 2.87e-03                         & \textbf{2.11}                    & 2.10                              & \textbf{0.58}                     & 1.15e-03                         & \textbf{0.58}                    & \textbf{0.58}                    \\
\begin{tabular}[c]{@{}c@{}}SHAPEffects\\ (0.2-08)\end{tabular}   & \textbf{1.70}                     & 3.67e-03                         & 1.71                             & \textbf{1.69}                     & \textbf{2.10}                     & 3.27e-03                         & \textbf{2.11}                    & 2.10                              & \textbf{0.58}                     & 1.31e-03                         & \textbf{0.58}                    & 0.57                             \\
\begin{tabular}[c]{@{}c@{}}SHAPEffects\\ (0.15-085)\end{tabular} & \textbf{1.70}                     & 2.34e-03                         & \textbf{1.70}                    & \textbf{1.69}                     & \textbf{2.10}                     & 2.35e-03                         & \textbf{2.11}                    & \textbf{2.09}                     & \textbf{0.58}                     & 9.40e-04                         & \textbf{0.58}                    & \textbf{0.58}                    \\
\begin{tabular}[c]{@{}c@{}}SHAPEffects\\ (0.1-0.9)\end{tabular}  & \textbf{1.70}                     & 2.34e-03                         & \textbf{1.70}                    & \textbf{1.69}                     & \textbf{2.10}                     & 2.35e-03                         & \textbf{2.11}                    & \textbf{2.09}                     & \textbf{0.58}                     & 9.40e-04                         & \textbf{0.58}                    & \textbf{0.58}                    \\
\begin{tabular}[c]{@{}c@{}}SHAPEffects\\ (0.05-095)\end{tabular} & 1.79                              & 4.44e-03                         & 1.80                             & 1.78                              & 2.23                              & 3.76e-03                         & 2.24                             & 2.22                              & 0.53                              & 1.60e-03                         & 0.53                             & 0.52                            
\end{tabular}}
\caption{Test results on the second case for the incremental shift context}
\label{cuadro:sintetico4}
\end{table}

\begin{table}[h]
\resizebox{\textwidth}{!}{
\begin{tabular}{c|llll|llll|llll}
\multirow{2}{*}{\textbf{Algorithm}}                              & \multicolumn{4}{c|}{\textbf{MAE}}                                                                                                           & \multicolumn{4}{c|}{\textbf{RMSE}}                                                                                                          & \multicolumn{4}{c}{\textbf{R}}                                                                                                             \\
                                                                 & \multicolumn{1}{c}{\textbf{Mean}} & \multicolumn{1}{c}{\textbf{Std}} & \multicolumn{1}{c}{\textbf{Max}} & \multicolumn{1}{c|}{\textbf{Min}} & \multicolumn{1}{c}{\textbf{Mean}} & \multicolumn{1}{c}{\textbf{Std}} & \multicolumn{1}{c}{\textbf{Max}} & \multicolumn{1}{c|}{\textbf{Min}} & \multicolumn{1}{c}{\textbf{Mean}} & \multicolumn{1}{c}{\textbf{Std}} & \multicolumn{1}{c}{\textbf{Max}} & \multicolumn{1}{c}{\textbf{Min}} \\ \hline
Powershap                                                        & \textbf{1.289}                    & 0.002                            & \textbf{1.293}                   & \textbf{1.285}                    & \textbf{1.583}                    & 0.002                            & \textbf{1.586}                   & \textbf{1.578}                    & \textbf{0.733}                    & 6.503e-04                        & \textbf{0.734}                   & \textbf{0.732}                   \\
Boruta-Shap                                                      & \textbf{1.289}                    & 0.002                            & \textbf{1.293}                   & \textbf{1.285}                    & \textbf{1.583}                    & 0.002                            & \textbf{1.586}                   & \textbf{1.578}                    & \textbf{0.733}                    & 6.503e-04                        & \textbf{0.734}                   & \textbf{0.732}                   \\
Shapicant                                                        & 1.291                             & 0.002                            & 1.295                            & 1.287                             & 1.587                             & 0.002                            & 1.591                            & 1.583                             & 0.731                             & 6.189e-04                        & 0.733                            & 0.730                            \\
Boruta                                                           & 1.291                             & 0.002                            & 1.295                            & 1.287                             & 1.587                             & 0.002                            & 1.591                            & 1.583                             & 0.731                             & 6.189e-04                        & 0.733                            & 0.730                            \\
PIMP                                                             & 1.351                             & 0.002                            & 1.357                            & 1.347                             & 1.669                             & 0.002                            & 1.673                            & 1.666                             & 0.703                             & 6.029e-04                        & 0.704                            & 0.702                            \\
\begin{tabular}[c]{@{}c@{}}Best Lasso \\ (0.001)\end{tabular}    & \textbf{1.289}                    & 0.002                            & \textbf{1.293}                   & \textbf{1.285}                    & \textbf{1.583}                    & 0.002                            & \textbf{1.586}                   & \textbf{1.578}                    & \textbf{0.733}                    & 6.503e-04                        & \textbf{0.734}                   & \textbf{0.732}                   \\
\begin{tabular}[c]{@{}c@{}}SHAPEffects\\ (0.25-075)\end{tabular} & 1.294                             & 0.002                            & 1.300                            & 1.290                             & 1.591                             & 0.002                            & 1.595                            & 1.587                             & 0.730                             & 6.990e-04                        & 0.731                            & 0.728                            \\
\begin{tabular}[c]{@{}c@{}}SHAPEffects\\ (0.2-08)\end{tabular}   & 1.295                             & 0.002                            & 1.301                            & 1.291                             & 1.593                             & 0.002                            & 1.598                            & 1.589                             & 0.729                             & 7.447e-04                        & 0.731                            & 0.727                            \\
\begin{tabular}[c]{@{}c@{}}SHAPEffects\\ (0.15-085)\end{tabular} & 1.291                             & 0.002                            & 1.295                            & 1.287                             & 1.587                             & 0.002                            & 1.591                            & 1.583                             & 0.731                             & 6.189e-04                        & 0.733                            & 0.730                            \\
\begin{tabular}[c]{@{}c@{}}SHAPEffects\\ (0.1-0.9)\end{tabular}  & 1.294                             & 0.002                            & 1.298                            & 1.290                             & 1.589                             & 0.002                            & 1.593                            & 1.585                             & 0.731                             & 7.544e-04                        & 0.732                            & 0.729                            \\
\begin{tabular}[c]{@{}c@{}}SHAPEffects\\ (0.05-095)\end{tabular} & 1.293                             & 0.002                            & 1.297                            & 1.286                             & 1.587                             & 0.002                            & 1.591                            & 1.581                             & 0.731                             & 6.873e-04                        & 0.733                            & 0.730                           
\end{tabular}}
\caption{Test results on the third case for the sudden shift context}
\label{cuadro:sintetico5}
\end{table}

\begin{table}[h]
\resizebox{\textwidth}{!}{
\begin{tabular}{c|llll|llll|llll}
\multirow{2}{*}{\textbf{Algorithm}}                              & \multicolumn{4}{c|}{\textbf{MAE}}                                                                                                           & \multicolumn{4}{c|}{\textbf{RMSE}}                                                                                                          & \multicolumn{4}{c}{\textbf{R}}                                                                                                             \\
                                                                 & \multicolumn{1}{c}{\textbf{Mean}} & \multicolumn{1}{c}{\textbf{Std}} & \multicolumn{1}{c}{\textbf{Max}} & \multicolumn{1}{c|}{\textbf{Min}} & \multicolumn{1}{c}{\textbf{Mean}} & \multicolumn{1}{c}{\textbf{Std}} & \multicolumn{1}{c}{\textbf{Max}} & \multicolumn{1}{c|}{\textbf{Min}} & \multicolumn{1}{c}{\textbf{Mean}} & \multicolumn{1}{c}{\textbf{Std}} & \multicolumn{1}{c}{\textbf{Max}} & \multicolumn{1}{c}{\textbf{Min}} \\ \hline
Powershap                                                        & 1.291                             & 0.002                            & 1.295                            & 1.286                             & 1.585                             & 0.002                            & 1.589                            & 1.581                             & 0.732                             & 7.101e-04                        & 0.733                            & 0.730                            \\
Boruta-Shap                                                      & \textbf{1.289}                    & 0.002                            & \textbf{1.294}                   & \textbf{1.284}                    & \textbf{1.583}                    & 0.002                            & \textbf{1.588}                   & \textbf{1.578}                    & \textbf{0.733}                    & 7.179e-04                        & \textbf{0.734}                   & \textbf{0.731}                   \\
Shapicant                                                        & 1.291                             & 0.002                            & 1.295                            & 1.288                             & 1.587                             & 0.002                            & 1.592                            & 1.584                             & 0.731                             & 6.103e-04                        & 0.732                            & 0.730                            \\
Boruta                                                           & 1.291                             & 0.002                            & 1.295                            & 1.288                             & 1.587                             & 0.002                            & 1.592                            & 1.584                             & 0.731                             & 6.103e-04                        & 0.732                            & 0.730                            \\
PIMP                                                             & 1.351                             & 0.002                            & 1.355                            & 1.347                             & 1.669                             & 0.002                            & 1.672                            & 1.665                             & 0.703                             & 5.940e-04                        & 0.704                            & 0.702                            \\
\begin{tabular}[c]{@{}c@{}}Best Lasso \\ (0.001)\end{tabular}    & 1.289                             & 0.002                            & \textbf{1.294}                   & \textbf{1.284}                    & \textbf{1.583}                    & 0.002                            & \textbf{1.588}                   & \textbf{1.578}                    & \textbf{0.733}                    & 7.179e-04                        & \textbf{0.734}                   & \textbf{0.731}                   \\
\begin{tabular}[c]{@{}c@{}}SHAPEffects\\ (0.25-075)\end{tabular} & 1.293                             & 0.002                            & 1.298                            & 1.287                             & 1.588                             & 0.002                            & 1.594                            & 1.582                             & 0.731                             & 7.552e-04                        & 0.733                            & 0.729                            \\
\begin{tabular}[c]{@{}c@{}}SHAPEffects\\ (0.2-08)\end{tabular}   & 1.294                             & 0.002                            & 1.299                            & 1.290                             & 1.591                             & 0.002                            & 1.596                            & 1.587                             & 0.730                             & 6.544e-04                        & 0.731                            & 0.728                            \\
\begin{tabular}[c]{@{}c@{}}SHAPEffects\\ (0.15-085)\end{tabular} & \textbf{1.289}                    & 0.002                            & \textbf{1.294}                   & \textbf{1.284}                    & \textbf{1.583}                    & 0.002                            & \textbf{1.588}                   & \textbf{1.578}                    & \textbf{0.733}                    & 7.179e-04                        & \textbf{0.734}                   & \textbf{0.731}                   \\
\begin{tabular}[c]{@{}c@{}}SHAPEffects\\ (0.1-0.9)\end{tabular}  & 1.294                             & 0.002                            & 1.300                            & 1.290                             & 1.589                             & 0.003                            & 1.595                            & 1.582                             & 0.730                             & 8.828e-04                        & 0.733                            & 0.728                            \\
\begin{tabular}[c]{@{}c@{}}SHAPEffects\\ (0.05-095)\end{tabular} & \textbf{1.289}                    & 0.002                            & \textbf{1.294}                   & \textbf{1.284}                    & \textbf{1.583}                    & 0.002                            & \textbf{1.588}                   & \textbf{1.578}                    & \textbf{0.733}                    & 7.179e-04                        & \textbf{0.734}                   & \textbf{0.731}                  
\end{tabular}}
\caption{Test results on the third case for the incremental shift context}
\label{cuadro:sintetico6}
\end{table}

\newpage

In the scenarios characterized by substantial or moderate influence of the variables, our proposed methodology clearly demonstrates its advantage, yielding superior results across all provided configurations. On the contrary, when the influence of the variables is minimal, the algorithms exhibit a similar behavior. Moreover, no significant distinction is observed between the results obtained in sudden shift cases and incremental shift cases, emphasizing the robustness of our algorithm.\\

In this synthetic example, we have exclusively examined a concept shift that affects the entire validation set. It is worth noting that if the concept shift were to be incorporated into the training set, the results across all methodologies could potentially improve, depending on the model's ability to effectively learn from this shift. Conversely, if the concept shift were to occur later and later in the validation data, at a certain point our algorithm would no longer eliminate the variables, resulting in comparable outcomes with the other methods, in a similar way to what happened in the previous examples with the incremental concept shift and small coefficient.

\subsubsection{Electricity Price Forecasting (EPF)}

\begin{figure}[h]
\centering
\includegraphics[scale=0.3]{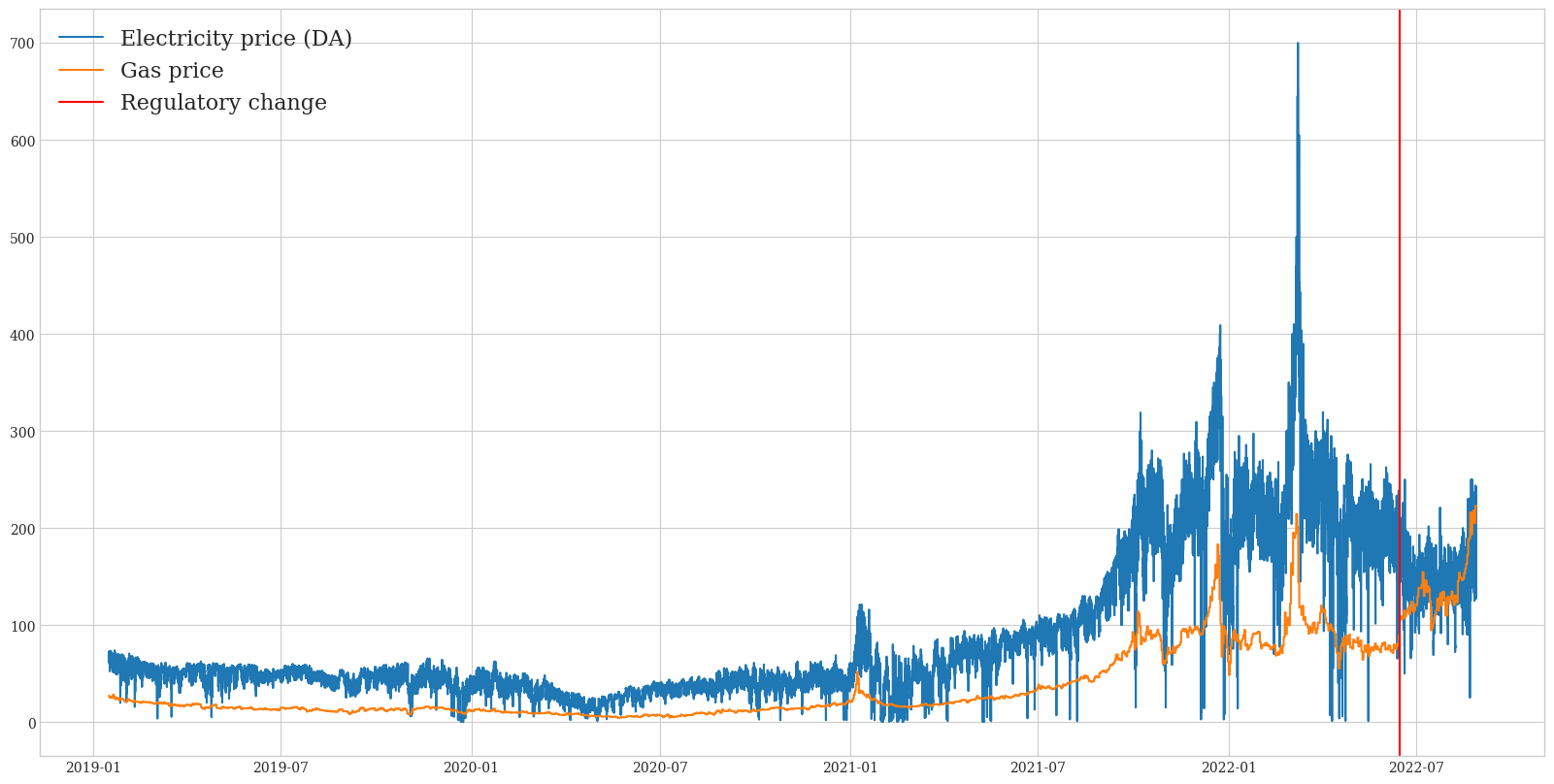}
\caption{The time series of the Day-Ahead market price and the gas price are observed. Before the regulatory change it can be seen that certain peaks in the price of electricity correspond to peaks in the price of gas. After the regulatory change the relationship is not so direct.}
\label{fig:da_vs_gas}
\end{figure}

\begin{figure}[h]
\centering
\includegraphics[scale=0.25]{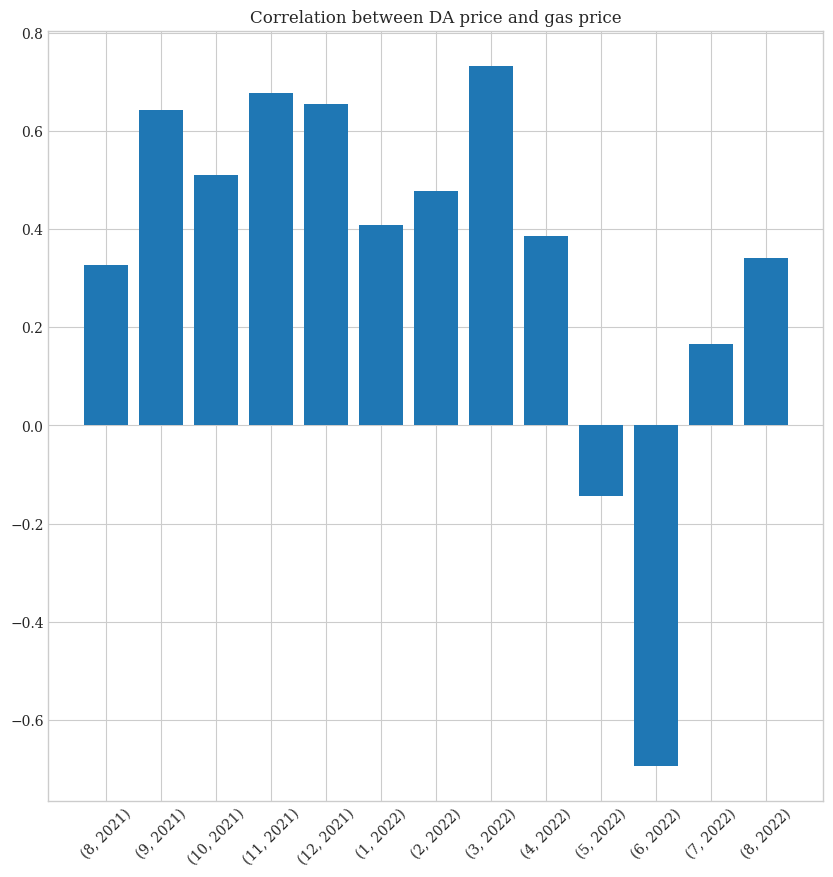}
\caption{The correlation between the electricity price and gas price variables by month can be seen: the relationship from May 2022 onwards has changed and, although it seems to recover, it should be taken into account that this indicator does not represent that the relationship is the same as before.}
\label{fig:correlation_da_vs_gas}
\end{figure}

The literature related to electricity price forecasting is extensive, with day-ahead (12 to 36 hours) forecasting prior to the Day-Ahead Market being the most studied case from a variety of perspectives \citep{maciejowska2022forecasting, lago2021forecasting, weron2014electricity}. However, the electricity market is constantly subject to regulatory changes and is highly influenced by political and economic situations. Specifically, in the Iberian market, given the crisis related to the gas price in mid-2022, a regulatory measure related to the gas price subvention to thermal power plants was established on June 15 2022 \cite{del2000ley}, changing the dynamics of the electricity market. Predictive models typically use the price of different fuels as regressors, among them the price of gas \citep{ortiz2016price, marcjasz2022distributional, shiri2015electricity}, so this change of behaviour has created a situation of uncertainty around the models. Figures \ref{fig:da_vs_gas} and \ref{fig:correlation_da_vs_gas} graphically show the change in the relationship between the variables. There are two reasons for studying this particular case: firstly, the issue at hand is considerably familiar to the authors, which makes it easier and safer to interpret. Secondly, we know that a concept shift situation exists and at least one of the variables that is causing it, which allows us to work in a controlled environment as well as with real data.\\

The question that arises in this case is whether keeping variables related with the gas price in the models is useful to predict the current situation. A dataset with 335 explanatory variables linked to the electricity market, several connected to the gas price, is available to predict the price of the Day-Ahead Market. The data considered is publicly available, mostly obtained from ESIOS portal, which is an information system developed by Red Eléctrica de España, and the data related to the gas price is obtained from the Iberian Gas Market (MIBGAS). The main variables considered are forecasts related to renewable energy generation, load forecasts, prices from the different markets, prices of the previous Daily Markets in France and lagged values of the series to be predicted. To describe the current behavior of the market with respect to each of the characteristics, statistics of some of these same variables during the last week are also included. These are the mean, standard deviation, median, minimum, maximum, first quartile, third quartile, skewness and kurtosis. The data start in January 2021 and end in August 2022, both included. The data are divided into: January 1, 2021 to May 31, 2022 for training, June 1, 2022 to July 31, 2022 as validation, and August 2022 as test.\\

\paragraph*{Results}Following the methodology described above, the results obtained are shown in Table \ref{cuadro:epf}.\\

\begin{table}[h]
\centering
\resizebox{\textwidth}{!}{
\begin{tabular}{c|cccc|cccc|cccc|c}
\multirow{2}{*}{\textbf{Algorithm}}                             & \multicolumn{4}{c|}{\textbf{MAE}}                                & \multicolumn{4}{c|}{\textbf{RMSE}}                              & \multicolumn{4}{c|}{\textbf{R}$^2$}                               & \multirow{2}{*}{\textbf{\begin{tabular}[c]{@{}c@{}}Number of\\ variables\end{tabular}}} \\
                                                                & \textbf{Mean}  & \textbf{Std.} & \textbf{Max}   & \textbf{Min}   & \textbf{Mean} & \textbf{Std.} & \textbf{Max}   & \textbf{Min}   & \textbf{Mean} & \textbf{Std.} & \textbf{Max}  & \textbf{Min}  &                                                                                         \\ \hline
Powershap                                                       & 23,13          & 2,23          & 31,88          & 19,28          & 29,37         & 2,39          & 38,76          & 25,25          & 0,09          & 0,16          & 0,33          & -0,57         & 107                                                                                     \\
Boruta-Shap                                                     & 23,2           & 2,06          & 30,57          & 19,65          & 29,58         & 2             & 36,14          & 25,79          & 0,08          & 0,13          & 0,31          & -0,36         & 53                                                                                      \\
Shapicant                                                       & 22,04          & 1,5           & 25,85          & 19,26          & 28,24         & 1,63          & 32,49          & 24,94          & 0,16          & 0,1           & 0,35          & -0,1          & 11                                                                                      \\
Boruta                                                          & 23,49          & 3,16          & 33,86          & 18,35          & 29,84         & 3,25          & 40,55          & 23,72          & 0,06          & 0,22          & 0,41          & -0,72         & 68                                                                                      \\
PIMP                                                            & 23,31          & 1,73 & 27,72          & 20,11          & 29,49         & 1,75          & 33,96          & 26,19          & 0,09          & 0,11          & 0,28          & -0,2         & 56                                                                                       \\
\begin{tabular}[c]{@{}c@{}}Best Lasso\\ (0.0001)\end{tabular}   & 22,79          & 1,81          & 30,01          & 19,5           & 29,22         & 1,96          & 36,03          & 25,96          & 0,1           & 0,13          & 0,3           & -0,36         & 47                                                                                      \\
\begin{tabular}[c]{@{}c@{}}SHAPEffects\\ 0.25-0.75\end{tabular} & 18,27          & 0,99          & 20,27          & 16,47          & 23,36         & 1,13          & 26,48          & 21,19          & 0,43          & 0,06          & 0,53          & 0,27          & 122                                                                                     \\
\begin{tabular}[c]{@{}c@{}}SHAPEffects\\ 0.2-0.8\end{tabular}   & \textbf{16,97} & 0,77          & \textbf{19,32} & \textbf{15,77} & \textbf{21,5} & 0,86          & \textbf{23,36} & \textbf{20,02} & 0,52          & \textbf{0,04} & \textbf{0,58} & \textbf{0,43} & 111                                                                                     \\
\begin{tabular}[c]{@{}c@{}}SHAPEffects\\ 0.15-0.85\end{tabular} & 19,01          & 1,69          & 23,88          & 16,4           & 23,78         & 1,83          & 28,83          & 21             & 0,41          & 0,09          & 0,54          & 0,13          & 136                                                                                     \\
\begin{tabular}[c]{@{}c@{}}SHAPEffects\\ 0.1-0.9\end{tabular}   & 20,57          & 1,88          & 27,72          & 16,7           & 25,77         & 2,1           & 33,51          & 21,51          & 0,3           & 0,12          & 0,52          & -0,17         & 78                                                                                      \\
\begin{tabular}[c]{@{}c@{}}SHAPEffects\\ 0.05-0.95\end{tabular} & 21             & 1,5           & 24,21          & 18,5           & 27,19         & 1,68          & 31,34          & 23,87          & 0,23          & 0,1           & 0,4           & -0,03         & 133                                                                                    
\end{tabular}}
\caption{Daily Market price prediction test results}
\label{cuadro:epf}
\end{table}

\begin{figure}[h]
\centering
\includegraphics[scale=0.3]{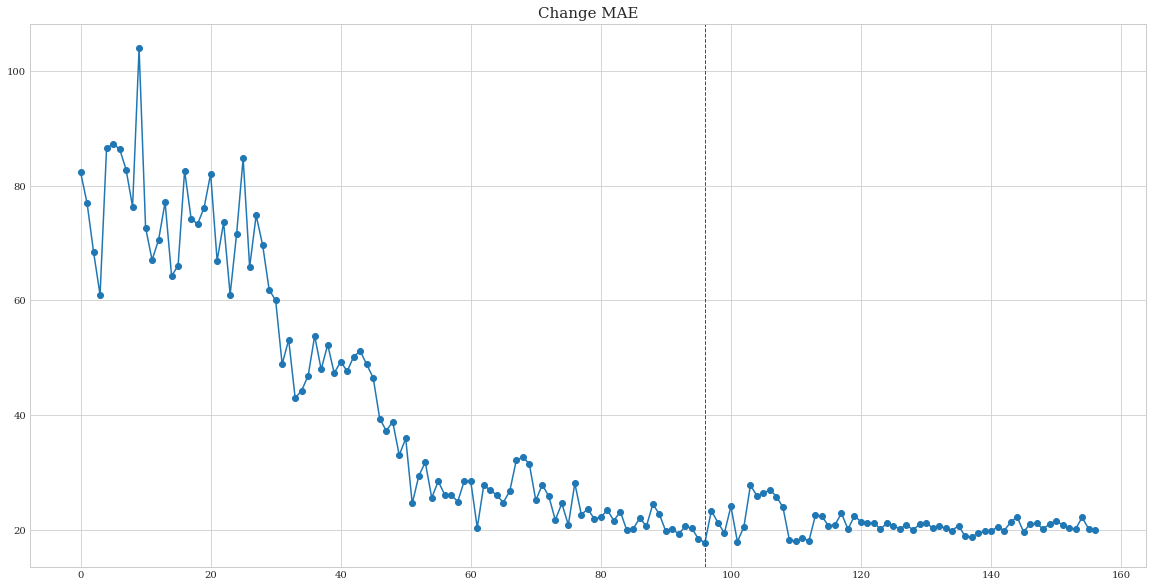}
\caption{Variation of MAE in the validation set over the iterations of the proposed variable selection procedure. The configuration shown is $(0.2, 0.8)$}
\label{fig:variacion_mae}
\end{figure}

The advantage of the proposed method is evident in all aspects. All the proposed configurations achieve better results than any other method analyzed. The configurations $(0.25, 0.75)$ and $(0.2, 0.8)$ eliminate all the variables related to the gas price, obtaining the best results. Moreover, in these two configurations the worst-case iterations (Max column for MAE and RMSE and Min column for R$^2$) are better than the average of the other variable selection methods. In addition, not only better results are obtained, but they are also more stable: the standard deviation is noticeably smaller in general than with the other algorithms. In particular, the best configuration is the one with the lowest variance. Furthermore, the percentage of improvement between our methodology and the other methodologies is around 20-25\%, which is significative. The number of variables selected seems to be higher than with the other methods; however, the improvement in the quality of the predictions justifies this selection. This essentially highlights the absence of overfitting concerns due to a high number of features selected in alternative methodologies, as our algorithm makes an even more extensive selection. The phenomenon observed here is that the features chosen by our method exhibit higher predictive performance within the current test set context, which is the actual key in feature selection. Recall that the selected variables are those that obtain the best MAE in the validation (Figure \ref{fig:variacion_mae}).\\

Another possible selection, as discussed above, would be the use of a smaller number of variables with a slightly higher MAE, since it can be observed that the elimination of the last variables does not produce a significant increase in MAE. It is also possible to observe a considerable increase in the MAE seven iterations later from the optimal MAE obtained during the process. It could be studied which variable causes this increase and if its elimination is really advantageous.

\subsubsection{Another real world example}
In order to validate the results, we analyzed one more real world data example: the Sberbank Russian Housing Market dataset \citep{sberbank-russian-housing-market}. It corresponds to data from a Kaggle competition with the objective of predicting the price of different houses. As the objective is not to obtain the best result in this dataset, only the training data have been used, which have been divided into three different sets chronologically, previously eliminating all the null data. As explanatory variables we have different variables related to information about the area in which each property is located. On the Kaggle competition website itself, a concept shift situation is described: `\textit{Although the housing market is relatively stable in Russia, the country’s volatile economy makes forecasting prices as a function of apartment characteristics a unique challenge}'. This perfectly describes a common situation in which we know that we are in a concept shift scenario, but  our knowledge is limited regarding the presence of any potentially responsible feature, and, if such a feature exists, which specific one it may be.

\paragraph*{Results}
The results are shown in Table \ref{cuadro:Sberbank}.\\

\begin{sidewaystable}
\centering
\resizebox{\textwidth}{!}{
\begin{tabular}{c|llll|llll|llll|c}
\multirow{2}{*}{\textbf{Algorithm}}                             & \multicolumn{4}{c|}{\textbf{MAE}}                                                                                                            & \multicolumn{4}{c|}{\textbf{RMSE}}                                                                                                           & \multicolumn{4}{c|}{\textbf{R$^2$}}                                                                                                              & \multirow{2}{*}{\textbf{\begin{tabular}[c]{@{}c@{}}Number of\\ variables\end{tabular}}} \\
                                                                & \multicolumn{1}{c}{\textbf{Mean}} & \multicolumn{1}{c}{\textbf{Std.}} & \multicolumn{1}{c}{\textbf{Max}} & \multicolumn{1}{c|}{\textbf{Min}} & \multicolumn{1}{c}{\textbf{Mean}} & \multicolumn{1}{c}{\textbf{Std.}} & \multicolumn{1}{c}{\textbf{Max}} & \multicolumn{1}{c|}{\textbf{Min}} & \multicolumn{1}{c}{\textbf{Mean}} & \multicolumn{1}{c}{\textbf{Std.}} & \multicolumn{1}{c}{\textbf{Max}} & \multicolumn{1}{c|}{\textbf{Min}} &                                                                                         \\ \hline
Powershap                                                       & 2,972E+06                         & 2,257E+04                         & 3,013E+06                        & 2,922E+06                         & 4,899E+06                         & 6,584E+04                         & 5,002E+06                        & 4,771E+06                         & 0,53                              & 1,300E-02                         & 0,55                             & 0,51                              & 7                                                                                       \\
Boruta-Shap                                                     & 2,977E+06                         & 1,747E+04                         & 3,008E+06                        & 2,942E+06                         & 4,907E+06                         & 6,457E+04                         & 5,055E+06                        & 4,778E+06                         & 0,53                              & 1,300E-02                         & 0,55                             & 0,5                               & 8                                                                                       \\
Shapicant                                                       & 2,658E+06                         & 3,425E+04                         & 2,738E+06                        & 2,571E+06                         & 4,357E+06                         & 7,911E+04                         & 4,593E+06                        & 4,232E+06                         & 0,62                              & 1,000E-02                         & 0,65                             & 0,58                              & 15                                                                                      \\
Boruta                                                          & 3,023E+06                         & 2,024E+04                         & 3,064E+06                        & 2,983E+06                         & 4,997E+06                         & 5,792E+04                         & 5,104E+06                        & 4,836E+06                         & 0,51                              & 1,100E-02                         & 0,54                             & 0,49                              & 6                                                                                       \\
PIMP                                                            & 2,672E+06                         & 3,595E+04                         & 2,737E+06                        & 2,580E+06                         & 4,428E+06                         & 8,025E+04                         & 4,628E+06                        & 4,265E+06                         & 0,61                              & 1,400E-02                         & 0,64                             & 0,58                              & 26                                                                                       \\
\begin{tabular}[c]{@{}c@{}}Best Lasso\\ (0.00001)\end{tabular}    & 2,610E+06                         & 2,831E+04                         & 2,671E+06                        & 2,560E+06                         & 4,292E+06                         & 7,824E+04                         & 4,465E+06                        & 4,127E+06                         & 0,64                              & 1,300E-02                         & 0,66                             & 0,61                              & 177                                                                                     \\
\begin{tabular}[c]{@{}c@{}}SHAPEffects\\ 0.25-0.75\end{tabular} & 2,615E+06                         & 3,273E+04                         & 2,677E+06                        & 2,517E+06                         & 4,311E+06                         & 8,586E+04                         & 4,505E+06                        & 4,119E+06                         & 0,63                              & 1,500E-02                         & 0,67                             & 0,6                               & 58                                                                                      \\
\begin{tabular}[c]{@{}c@{}}SHAPEffects\\ 0.2-0.8\end{tabular}   & 2,602E+06                         & 2,612E+04                         & 2,662E+06                        & 2,557E+06                         & 4,301E+06                         & 7,870E+04                         & 4,506E+06                        & 4,137E+06                         & 0,64                              & 1,300E-02                         & 0,66                             & 0,6                               & 39                                                                                      \\
\begin{tabular}[c]{@{}c@{}}SHAPEffects\\ 0.15-0.85\end{tabular} & 2,593E+06                         & 3,495E+04                         & 2,692E+06                        & 2,524E+06                         & 4,268E+06                         & 8,643E+04                         & 4,472E+06                        & 4,123E+06                         & 0,64                              & 1,500E-02                         & 0,67                             & 0,61                              & 67                                                                                      \\
\begin{tabular}[c]{@{}c@{}}SHAPEffects\\ 0.1-0.9\end{tabular}   & \textbf{2,567E+06}                & 2,219E+04                         & \textbf{2,604E+06}               & \textbf{2,508E+06}                & \textbf{4,229E+06}                & 7,021E+04                         & \textbf{4,374E+06}               & \textbf{4,011E+06}                & \textbf{0,65}                     & 1,200E-02                         & \textbf{0,68}                    & \textbf{0,62}                     & 35                                                                                      \\
\begin{tabular}[c]{@{}c@{}}SHAPEffects\\ 0.05-0.95\end{tabular} & 2,639E+06                         & 2,730E+04                         & 2,718E+06                        & 2,574E+06                         & 4,405E+06                         & 8,783E+04                         & 4,575E+06                        & 4,186E+06                         & 0,62                              & 1,500E-02                         & 0,66                             & 0,59                              & 123                                                                                    
\end{tabular}
}
\caption{Test results on the Sberbank Russian Housing Market dataset\protect\footnotemark}
\footnotetext[1]{The preprocessing phase was not applied to this dataset because it removed many variables. (Section \ref{sec:algoritmo})}
\label{cuadro:Sberbank}
\end{sidewaystable}

Once more, a notable disparity becomes apparent between each configuration of the proposed method and the remaining algorithms, with the former standing out prominently. The algorithm that achieves closer results is the Lasso, but uses five times more variables than the best provided configuration so, in this case, this extended selection is not justified. Again, in this instance, the worst results obtained in each metric by the best configuration of our methodology is better than the best results obtained by most of the other algorithms. Moreover, in this dataset there is up to a 14\% improvement, which is considerable but not as high as in the EPF example. In this particular instance, the specific variables that were excluded, resulting in the algorithm's superior performance, remain unidentified, but it would be interesting to analyze each one of the features that produces big decreases in MAE during the procedure. As mentioned earlier, it is important to note that the intention of the proposed algorithm is not to detect these features but rather to eliminate them if they are present.

\subsection{Static scenarios}

Other datasets presenting a more static situation between training, validation and test sets are now analyzed. It is also relevant to study the behaviour in this context, because we would like the algorithm to behave as good as possible in all kinds of scenarios. They are all available in Kaggle or in the UCI Machine Learning repository \citep{Dua:2019}:

\begin{itemize}
\item \textbf{CAT Scan Localization\footnote{\url{https://archive.ics.uci.edu/ml/datasets/Relative+location+of+CT+slices+on+axial+axis}}}. This dataset consists of 384 features extracted from Computed Tomography (CT) images. The target variable denotes the relative location of the CT slice on the axial axis of the human body. The data were obtained from a set of 53500 CT images from 74 different patients. Each CT slice is described by two histograms in polar space. The first histogram describes the location of bony structures in the image, the second one the location of air inclusions inside the body. Both histograms are concatenated to form the final feature vector. Bins outside the image are marked with the value -0.25. The target variable (relative location of an image on the axial axis) was constructed by manually annotating up to 10 different landmarks in each CT volume with known location. The location of the slices between landmarks was interpolated. The division into training, validation and test is done randomly.\\

\item \textbf{Appliances Energy Prediction\footnote{\url{https://archive.ics.uci.edu/ml/datasets/Appliances+energy+prediction}}} \citep{candanedo2017data}. The aim in this dataset is to predict the energy consumption of household appliances in a low energy building. For this purpose, the consumption of the appliances every 10 minutes has been stored for 4 and a half months. There are 28 possible explanatory variables, including temperature, humidity and different weather conditions of nearby places of interest. There are also two random variables in the dataset. The division into training, validation and test is random.\\

\item \textbf{Max Planck Weather Dataset \footnote{\url{https://www.bgc-jena.mpg.de/wetter/}}}. It is a dataset with 12 atmospheric characteristics over time taken every 10 minutes. The variable to predict is the wind speed after first filtering only by hourly values and eliminating data with negative wind speed. In order to have more variables, lags from one day to one week are considered for each of the variables, resulting in 80 possible explanatory variables, also considering month and time. The division is also done chronologically: from 2009 to 2015 for training, 2015 for validation and 2016 for testing.
\end{itemize}

\paragraph*{Results}All results are shown in the Tables \ref{cuadro:ct} - \ref{cuadro:weather}.\\

In general, very close results can be observed among all methods. For the CAT Scan Localization dataset (Table \ref{cuadro:ct}), the selection of variables through Lasso regression seems to obtain the best results. The proposed method behaves similarly to the other algorithms, with very similar results for all five configurations. A similar situation occurs with the Appliances Energy Prediction dataset (Table \ref{cuadro:appliances}), where Powershap, Boruta and Boruta-Shap obtain the best result. The proposed method is slightly worse than them and better than the other methods. In the remaining dataset (Table \ref{cuadro:weather}), the described methodology obtains the best results, selection via Lasso regression performs similarly but with a much smaller number of variables. Thus, our methodology got equivalent results to the other methods in the three datasets, with a number of variables in the upper range of those of the models analyzed. In these instances, the variation in improvement between the different methodologies is only around 1\%, so we are able to say that the methods achieve equivalent results. Consequently, it can be asserted that the proposed methodology exhibits the capability to discern features that lack substantive impact on the target variable, thereby mitigating overfitting and the imposition of artificial relationships that are not actually present as well as the other methods.

\begin{table}[h]
\centering
\resizebox{\textwidth}{!}{
\begin{tabular}{c|cccc|cccc|cccc|c}
\multirow{2}{*}{\textbf{Algorithm}}                             & \multicolumn{4}{c|}{\textbf{MAE}}                           & \multicolumn{4}{c|}{\textbf{RMSE}}                          & \multicolumn{4}{c|}{\textbf{R$^2$}}                             & \multirow{2}{*}{\textbf{\begin{tabular}[c]{@{}c@{}}Number of\\ variables\end{tabular}}} \\
                                                                & \textbf{Mean} & \textbf{Std.} & \textbf{Max} & \textbf{Min} & \textbf{Mean} & \textbf{Std.} & \textbf{Max} & \textbf{Min} & \textbf{Mean} & \textbf{Std.} & \textbf{Max} & \textbf{Min} &                                                                                         \\ \hline
Powershap                                                       & 4,42          & 0,09          & 4,62         & 4,2          & 7,12          & 0,17          & 7,46         & 6,78         & \textbf{0,89}          & 5,00E-03      & \textbf{0,9}          & \textbf{0,88}         & 147                                                                                     \\
Boruta-Shap                                                     & 4,37          & 0,09          & 4,61         & 4,16         & 7,11          & 0,17          & 7,5          & 6,72         & \textbf{0,89}          & 5,00E-03      & \textbf{0,9}          & \textbf{0,88}         & 155                                                                                     \\
Shapicant                                                       & 4,79          & 0,07          & 4,92         & 4,59         & 8,07          & 0,14          & 8,34         & 7,7          & 0,86          & 5,00E-03      & 0,87         & 0,85         & 63                                                                                      \\
Boruta                                                          & 4,41          & 0,11          & 4,67         & \textbf{4,12}         & 7,19          & 0,19          & 7,55         & 6,62         & \textbf{0,89}          & 6,00E-03      & \textbf{0,9}          & 0,87         & 182                                                                                     \\
PIMP                                                            & 4,82         & 0,08          & 4,97        & 4,64        & 7,94         & 0,16      & 8,33        & 7,61        & 0,86          & 6,00E-03      & 0,87         & 0,85         & 33                                                                                       \\
\begin{tabular}[c]{@{}c@{}}Best Lasso\\ (0.001)\end{tabular}    & \textbf{4,32}          & 0,08          & \textbf{4,47}         & \textbf{4,12}         & \textbf{7,01}          & 0,15          & \textbf{7,31}         & \textbf{6,54}         & \textbf{0,89}          & 5,00E-03      & \textbf{0,9}          & \textbf{0,88}         & 98                                                                                      \\
\begin{tabular}[c]{@{}c@{}}SHAPEffects\\ 0.25-0.75\end{tabular} & 4,42          & 0,09          & 4,63         & 4,2          & 7,1           & 0,16          & 7,44         & 6,6          & \textbf{0,89}          & 5,00E-03      & \textbf{0,9}          & \textbf{0,88}         & 174                                                                                     \\
\begin{tabular}[c]{@{}c@{}}SHAPEffects\\ 0.2-0.8\end{tabular}   & 4,45          & 0,08          & 4,62         & 4,26         & 7,32          & 0,15          & 7,61         & 6,96         & 0,88          & 4,00E-03      & 0,89         & 0,87         & 159                                                                                     \\
\begin{tabular}[c]{@{}c@{}}SHAPEffects\\ 0.15-0.85\end{tabular} & 4,78          & 0,08          & 5,02         & 4,63         & 7,87          & 0,18          & 8,38         & 7,5          & 0,86          & 6,00E-03      & 0,88         & 0,84         & 105                                                                                     \\
\begin{tabular}[c]{@{}c@{}}SHAPEffects\\ 0.1-0.9\end{tabular}   & 4,45          & 0,11          & 4,72         & 4,2          & 7,23          & 0,18          & 7,78         & 6,66         & 0,88          & 5,00E-03      & \textbf{0,9}          & 0,87         & 158                                                                                     \\
\begin{tabular}[c]{@{}c@{}}SHAPEffects\\ 0.05-0.95\end{tabular} & 4,36          & 0,08          & 4,61         & 4,21         & 7,15          & 0,16          & 7,59         & 6,76         & \textbf{0,89}          & 5,00E-03      & \textbf{0,9}          & 0,87         & 168                                                                                    
\end{tabular}
}
\caption{Test results on CAT Scan Localization dataset}
\label{cuadro:ct}
\end{table}

\begin{table}[h]
\centering
\resizebox{\textwidth}{!}{
\begin{tabular}{c|cccc|cccc|cccc|c}
\multirow{2}{*}{\textbf{Algorithm}}                             & \multicolumn{4}{c|}{\textbf{MAE}}                               & \multicolumn{4}{c|}{\textbf{RMSE}}                               & \multicolumn{4}{c|}{\textbf{R$^2$}}                              & \multirow{2}{*}{\textbf{\begin{tabular}[c]{@{}c@{}}Number of\\ variables\end{tabular}}} \\
                                                                & \textbf{Mean}  & \textbf{Std.} & \textbf{Max}   & \textbf{Min}  & \textbf{Mean}  & \textbf{Std.} & \textbf{Max}   & \textbf{Min}   & \textbf{Mean} & \textbf{Std.} & \textbf{Max} & \textbf{Min}  &                                                                                         \\ \hline
Powershap                                                       & \textbf{42,24} & 0,23          & \textbf{42,74} & \textbf{41,7} & \textbf{81,84} & 0,41          & 82,79          & \textbf{81,02} & \textbf{0,39} & 6,00E-03      & \textbf{0,4} & 0,37          & 24                                                                                      \\
Boruta-Shap                                                     & \textbf{42,24} & 0,23          & \textbf{42,74} & \textbf{41,7} & \textbf{81,84} & 0,41          & 82,79          & \textbf{81,02} & \textbf{0,39} & 6,00E-03      & \textbf{0,4} & 0,37          & 24                                                                                      \\
Shapicant                                                       & 45,04          & 0,22          & 45,51          & 44,66         & 84,74          & 0,33          & 85,57          & 84             & 0,34          & 5,00E-03      & 0,35         & 0,33          & 10                                                                                      \\
Boruta                                                          & \textbf{42,24} & 0,23          & \textbf{42,74} & \textbf{41,7} & \textbf{81,84} & 0,41          & 82,79          & \textbf{81,02} & \textbf{0,39} & 6,00E-03      & \textbf{0,4} & 0,37          & 24                                                                                      \\
PIMP                                                            & 45,15          & 0,21          & 45,52          & 44,56          & 85,53          & 0,33          & 86,22          & 84,28          & 0,33          & 5,00E-03      & 0,35            & 0,32          & 10                                                                                      \\
\begin{tabular}[c]{@{}c@{}}Best Lasso\\ (0.00001)\end{tabular}    & 42,7           & 0,25          & 43,23          & 42,21         & 82,33          & 0,36          & 83,28          & 81,55          & 0,38          & 5,00E-03      & 0,39         & 0,37          & 26                                                                                      \\
\begin{tabular}[c]{@{}c@{}}SHAPEffects\\ 0.25-0.75\end{tabular} & 42,57          & 0,19          & 43,01          & 42,09         & 82,61          & 0,34          & 83,38          & 81,97          & 0,38          & 5,00E-03      & 0,39         & 0,36          & 23                                                                                      \\
\begin{tabular}[c]{@{}c@{}}SHAPEffects\\ 0.2-0.8\end{tabular}   & 42,57          & 0,19          & 43,01          & 42,09         & 82,61          & 0,34          & 83,38          & 81,97          & 0,38          & 5,00E-03      & 0,39         & 0,36          & 23                                                                                      \\
\begin{tabular}[c]{@{}c@{}}SHAPEffects\\ 0.15-0.85\end{tabular} & 42,57          & 0,19          & 43,01          & 42,09         & 82,61          & 0,34          & 83,38          & 81,97          & 0,38          & 5,00E-03      & 0,39         & 0,36          & 23                                                                                      \\
\begin{tabular}[c]{@{}c@{}}SHAPEffects\\ 0.1-0.9\end{tabular}   & 42,36          & 0,21          & 42,78          & 41,8          & 81,94          & 0,35          & \textbf{82,59} & 81,2           & \textbf{0,39} & 5,00E-03      & \textbf{0,4} & \textbf{0,38} & 25                                                                                      \\
\begin{tabular}[c]{@{}c@{}}SHAPEffects\\ 0.05-0.95\end{tabular} & 43,22          & 0,19          & 43,8           & 42,82         & 82,94          & 0,41          & 84,04          & 82             & 0,37          & 6,00E-03      & 0,38         & 0,35          & 19                                                                                     
\end{tabular}
}
\caption{Test results on Appliances Energy Prediction dataset}
\label{cuadro:appliances}
\end{table}

\begin{table}[h]
\centering
\resizebox{\textwidth}{!}{
\begin{tabular}{c|cccc|cccc|cccc|c}
\multirow{2}{*}{\textbf{Algorithm}}                             & \multicolumn{4}{c|}{\textbf{MAE}}                                & \multicolumn{4}{c|}{\textbf{RMSE}}                               & \multicolumn{4}{c|}{\textbf{R$^2$}}                                  & \multirow{2}{*}{\textbf{\begin{tabular}[c]{@{}c@{}}Number of\\ variables\end{tabular}}} \\
                                                                & \textbf{Mean}  & \textbf{Std.} & \textbf{Max}   & \textbf{Min}   & \textbf{Mean}  & \textbf{Std.} & \textbf{Max}   & \textbf{Min}   & \textbf{Mean}  & \textbf{Std.} & \textbf{Max}   & \textbf{Min}   &                                                                                         \\ \hline
Powershap                                                       & 1,044          & 0,003         & 1,053          & 1,037          & 1,402          & 0,004         & 1,41           & 1,395          & 0,168          & 0,004         & 0,176          & 0,158          & 79                                                                                      \\
Boruta-Shap                                                     & 1,044          & 0,003         & 1,051          & 1,038          & 1,402          & 0,003         & 1,408          & 1,398          & 0,168          & 0,003         & 0,173          & 0,161          & 33                                                                                      \\
Shapicant                                                       & 1,049          & 0,003         & 1,054          & 1,041          & 1,407          & 0,002         & 1,412          & 1,403          & 0,161          & 0,003         & 0,167          & 0,156          & 13                                                                                      \\
Boruta                                                          & 1,044          & 0,003         & 1,051          & 1,037          & 1,401          & 0,003         & 1,407          & 1,395          & 0,169          & 0,003         & 0,176          & 0,162          & 48                                                                                      \\
PIMP                                                            & 1,045          & 0,004        & 1,061          & 1,04          & 1,406          & 0,007        & 1,436          & 1,397          & 0,163          & 0,009        & 0,174          & 0,327          & 18                                                                                       \\
\begin{tabular}[c]{@{}c@{}}Best Lasso\\ (0.001)\end{tabular}    & 1,042          & 0,001         & 1,046          & 1,039          & \textbf{1,394} & 0,001         & \textbf{1,397} & 1,391          & \textbf{0,177} & 0,002         & 0,18           & \textbf{0,174} & 15                                                                                      \\
\begin{tabular}[c]{@{}c@{}}SHAPEffects\\ 0.25-0.75\end{tabular} & \textbf{1,039} & 0,002         & \textbf{1,045} & \textbf{1,034} & \textbf{1,394} & 0,003         & 1,4            & \textbf{1,389} & \textbf{0,177} & 0,003         & \textbf{0,184} & 0,17           & 75                                                                                      \\
\begin{tabular}[c]{@{}c@{}}SHAPEffects\\ 0.2-0.8\end{tabular}   & 1,042          & 0,003         & 1,054          & 1,037          & 1,399          & 0,004         & 1,413          & 1,39           & 0,172          & 0,005         & 0,182          & 0,155          & 74                                                                                      \\
\begin{tabular}[c]{@{}c@{}}SHAPEffects\\ 0.15-0.85\end{tabular} & 1,042          & 0,003         & 1,051          & 1,038          & 1,399          & 0,003         & 1,411          & 1,393          & 0,171          & 0,004         & 0,179          & 0,157          & 79                                                                                      \\
\begin{tabular}[c]{@{}c@{}}SHAPEffects\\ 0.1-0.9\end{tabular}   & 1,057          & 0,003         & 1,062          & 1,049          & 1,416          & 0,004         & 1,428          & 1,406          & 0,151          & 0,005         & 0,163          & 0,137          & 70                                                                                      \\
\begin{tabular}[c]{@{}c@{}}SHAPEffects\\ 0.05-0.95\end{tabular} & 1,043          & 0,003         & 1,051          & 1,037          & 1,398          & 0,003         & 1,407          & 1,393          & 0,173          & 0,003         & 0,179          & 0,162          & 63                                                                                     
\end{tabular}
}
\caption{Test results in the Max Planck Weather Dataset}
\label{cuadro:weather}
\end{table}
\renewcommand{\thefootnote}{\alph{footnote}}
\setcounter{footnote}{0}

\newpage

\section{Conclusions and future lines of research}\label{sec:conclusiones}

A new feature selection method for regression problems has been developed. The algorithm is based on the idea of observing how the variables of a certain model influence when making predictions, independently of the global influence they have. Specifically, a relationship is established between the model errors and the Shapley values, allowing for a more local analysis than the other feature selection algorithms. On the one hand, in situations where there is a concept shift, the algorithm is able to find the features that undergo this change of behavior in case they are available and have a negative influence on the predictions made by the model. Therefore, these variables are eliminated, leading to higher predictive performance and easier maintenance of the model after it has been put into production. On the other hand, in static situations, the model is able to detect the variables that cause overfitting obtaining comparable results with the state of the art. These variables create forced relationships in training, hence the effects they produce in the validation stage result in an undesired negative influence and, thus, are eliminated.\\

There are two clear problems to tackle as a continuation of this work. An interesting future line of research would be the omission of the parameters $q_{low}$ and $q_{high}$. As observed, these parameters of the algorithm can change the development of the algorithm iterations, producing different results in each case. Automatic selection of such quantiles in advance or varying between iterations in an optimal way would facilitate the use of the method. From a broader perspective, the generalization of our methodology to classification problems would be a natural extension of this research, contributing to the different studies that already exist on concept shift and feature selection in this area but with a different perspective.\\

\section*{Declarations}
\textbf{Acknowledgments} We thank Jesús Juan Ruiz for his insight into the problem and help. \\

\textbf{Funding} This work has been funded by grant MIG-20211033 from Centro para el Desarrollo Tecnológico Industrial, Ministerio de Universidades, and European Union-NextGenerationEU.\\

\textbf{Conflict of interest} The authors have nothing to declare.\\

\textbf{Ethics approval} Not applicable.\\

\textbf{Consent to participate} Not applicable.\\

\textbf{Consent for publication} Not applicable.\\

\textbf{Availability of data and material} All data is available in the links provided in the text.\\

\textbf{Code availability} All code is available in the link provided in \url{https://github.com/CCaribe9/SHAPEffects}\\

\textbf{Authors' contributions} All authors contributed to the study conception and design, methodology, analysis, investigation and writing of the manuscript. The full code was written by Carlos Sebastián.

\bibliography{references}

\end{document}